%% file: main.tex
\documentclass[conference]{IEEEtran}
\usepackage{times}

\usepackage[numbers]{natbib}
\usepackage{multicol}
\usepackage[bookmarks=true]{hyperref}
\usepackage{booktabs}
\usepackage{amsmath}
\usepackage{color}
\usepackage{changepage}
\usepackage{graphicx}
\usepackage{booktabs} 
\usepackage{makecell} 
\usepackage{float}

\begin{document}

\title{Learning Inverse Kinodynamics for \\ Autonomous Vehicle Drifting}
\author{
    Mihir Suvarna \\ msuvarna@cs.utexas.edu \\ {University of Texas at Austin} 
    \and Omeed Tehrani \\ omeed@cs.utexas.edu \\ {University of Texas at Austin} 
}

\maketitle

\IEEEpeerreviewmaketitle

\input{abstract.tex}

\input{intro.tex}

\input{related_work.tex}

\input{method.tex}

\input{experiments.tex}

\input{discussion.tex}

\input{conclusion.tex}

\input{acknowledgement.tex}

\bibliographystyle{plainnat}
\bibliography{references}

\end{document}

%% file: abstract.tex
\begin{abstract}
In this work, we explore a data-driven learning-based approach to learning the kinodynamic model of a small autonomous vehicle, and observe the effect it has on motion planning, specifically autonomous drifting. When executing a motion plan in the real world, there are numerous causes for error, and what is planned is often not what is executed on the actual car. Learning a kinodynamic planner based off of inertial measurements and executed commands can help us learn the world state. In our case, we look towards the realm of drifting; it is a complex maneuver that requires a smooth enough surface, high enough speed, and a drastic change in velocity. We attempt to learn the kinodynamic model for these drifting maneuvers, and attempt to tighten the slip of the car. Our approach is able to learn a kinodynamic model for high-speed circular navigation, and is able to avoid obstacles on an autonomous drift at high speed by correcting an executed curvature for loose drifts. We seek to adjust our kinodynamic model for success in tighter drifts in future work.
\end{abstract}

%% file: intro.tex
\section{Introduction}

\begin{figure}[h!]
  \centering
\includegraphics[width=.42\textwidth]{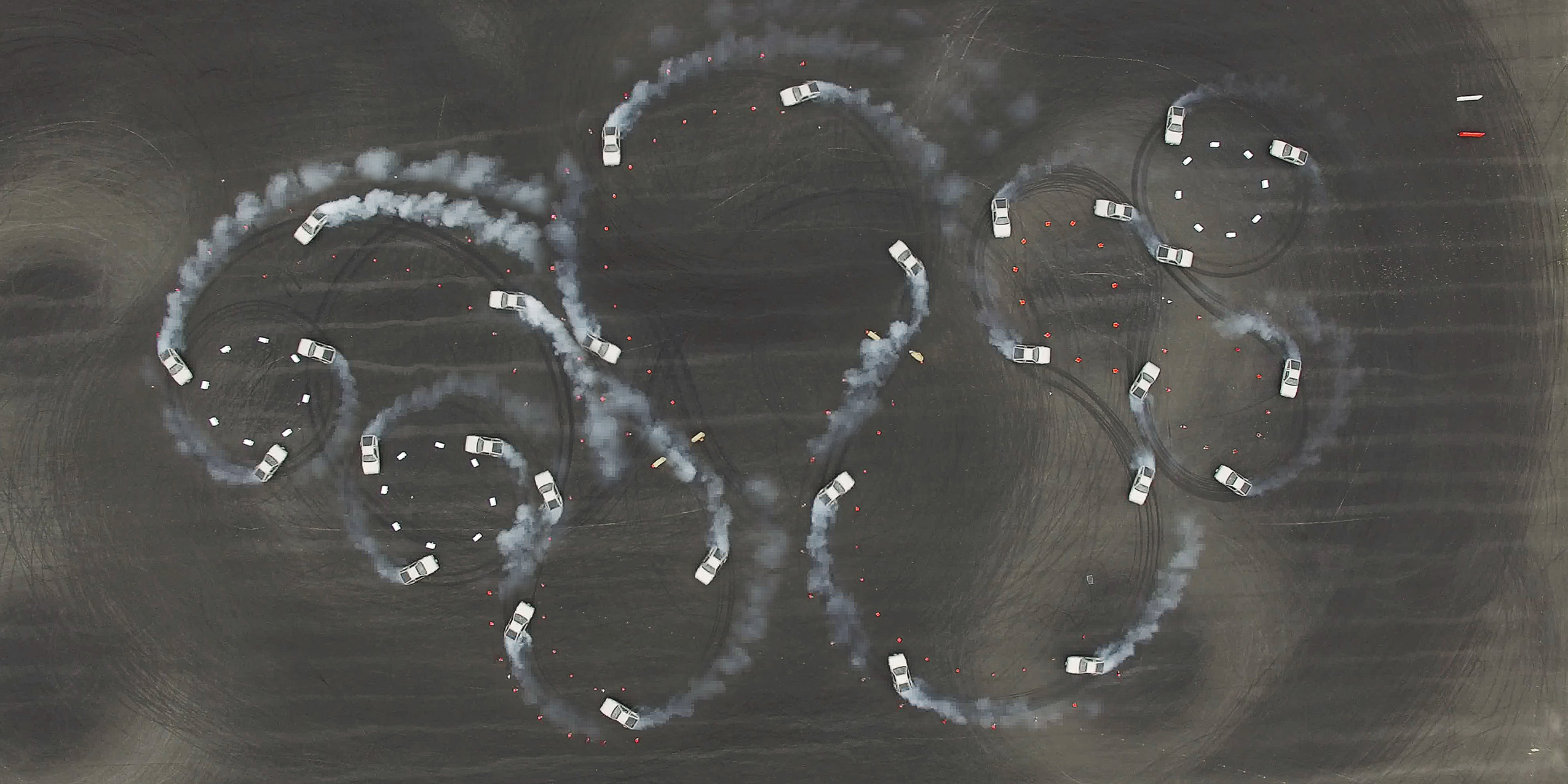}
\caption {Stanford's MARTYkhana DeLorean drifting on an open space and tracing its path.}
  \label{fig: marty}
\end{figure}

Navigation using task and motion planners for small-scale autonomous robots has been studied extensively (citation here), and explored in many different dimensions. However, motion planning for drifting is a much-less explored topic, and a key area of interest for this paper. Researchers have previously explored how to perform drifts in structured environments, including MARTYkhana (see Fig. \ref{fig: marty}) or even the autonomous drifting Supra from Toyota research. Drifting a full-scale vehicle requires a delicate balance between the speed of the car, the angle of the turn, and the traction of the tires. A driver must use the throttle, steering, and brakes to control the car's trajectory and maintain the slide through the turn. Shifting into focus for our project, we look directly at F1/10 vehicles, and performing drifts on these small-scale cars. 

\par Motion planning in such a domain is in itself a bit of a challenge, as the terrain that the plan is executed on can have a drastic impact on the actual impact of the plan. Thus, we set out to learn an inverse kinodynamic (IKD) model that would be able to use a data-driven approach from inertial measurements off the car and mappings from a tele-operated joystick. Existing approaches have learned the kinodynamic model for high-speed operations on different terrains, and we based our research off these foundational blocks and collected data in a much more hands-on fashion.

\par Essentially, a car's teleoperated sequence of events is similar to a driver actually controlling the car, commanding it at a given forward velocity and curvature. However, as mentioned earlier, this commanded sequence of controls is seldom executed in real life, notably the curvature. When feeding in a sequence of events to a small F1/10 car, a noticeable delta can be seen in the executed versus actual curvature, which we seek to correct in this paper. By collecting inertial measurements and teleoperated control data, we can learn an inverse kinodynamic model that can correct executed commands to be actual executions on the real vehicle. Our approach uses the commanded velocity and calculated angular velocity to create a mapping of the controls with the world state, and spits out a corrected angular velocity that is an overcorrection of the execution to get as close to the real-world execution as possible. We show that our IKD model gets very close to predicting the actual inertial data, and show concrete measurements of trajectories that prove this.

\par In research, the question has arose several times whether we can use drifting mechanics (autonomously) to prevent car crashes in the most extreme circumstances. In a sense, there may be some optimal sequence that can use telemetry and sensors to operate a different control sequence that avoids a head-on collision in an unfortunate event. We provide some future direction in lieu of this, and conduct some experiments in which we show that the slip of a car can be tightened by learning the inverse kinodynamic model, preventing a collision at a high speed drift.

%% file: related_work.tex
\section{Background and Related Work}
\label{sec:relatedwork}
In this section, we will discuss some previous works in lieu of the research that we are doing.
\subsection{Motion Planning for Mobile Robots}
The field of motion planning is vast within the realm of robotics, and explored heavily specifically within the field of mobile robots. Similar to the F1/10 vehicles that we work on in this paper, there are motion planning algorithms that exist, accounting for various factors. Some notable constraints are terrain, uncertain environments, and even actuation/latency issues when converting motion plans from simulation to real-world \cite{surveyMP}.

\par A large area of motion planning is hindered by the level of uncertainty in the real world. Previous research has also explored using reinforcement learning in the presence of uncertainty \cite{mprl} for autonomous vehicles. This line of research focused on factors related to obstacles while driving, and took on a learning-based approach to acting on the road. Usage of perception was a huge aid that was discussed \cite{mprl}, namely occlusions when cornering turns and how to approach such uncertainty. Our line of work in this paper does not account for perception, but it is a good future work proposal to integrate our motion-planning with perception to better model our IKD training and testing.

\subsection{Drifting for Vehicles}
Previous research has looked at research in the area of drifting, namely wheel slipping \cite{wheelslip}. Using inertial data, we can collect the actual executed data at the time of the drift, specifically where the car began to slip. This inertial measurement doesn't give much detail about the terrain, but noticing a sharp change in velocity and a sharp curvature is indicative of oversteering the vehicle. It is also important to note that drifting on a certain terrain might not always be replicable; previous research suggests that drifting is a very complex maneuver that often requires wheel slip to be executed perfectly, in which the car itself regains traction after the drift. 
\par Our work takes into account the inertial measurements of the car before, during, and after the drift, replicated across many trials to truly understand what the executed control mapped to the slippage of the wheels. More notably, we wish to correct the oversteer and instead understeer the car on a drift, preventing it from colliding with any unintentional objects. Previous research has explored correcting a drift after a rear-end collision, similar to our line of work \cite{driftcontrol}.

\subsection{Learning Inverse Kinodynamics}

Arguably the most important paper in regards to our work, the research in \cite{ikdhighsped} leverages a data driven approach with on-board inertial observations to learn an inverse kinodynamic model for accurate high-speed off road navigation on an unstructured terrain. In this paper, they leverage a learned inverse kinodynamic model to produce a system control input for every-time step, which is one of the key ideas that we take away from the paper in our work. The goal, generally, is to learn the function $f_{\theta}^{+}$ given the onboard inertial observations. More specifically, the paper formulates the function below: 

$$f_{\theta}^{+}(\Delta{x}, x, y) \approx f^{-1}(\Delta{x}, x, y)$$

\par The idea is that with enough data, the impact of the world state on the control of the vehicle becomes more predictable given observations related to the desired domain, which in the papers case was high speed, terrain aware navigation. Additionally, this paper leverages the same vehicle that we use for our work (the UT AUTOmata vehicle), is tested on several indoor and outdoor environments, and effectively outperforms the baseline model. We follow a simple formulation of these ideas in our work, which can be found in section \ref{method}.

%% file: method.tex
\section{Method} \label{method}

\subsection{Hardware}
\begin{figure}[h!]
  \centering
  \includegraphics[width=0.42\textwidth]{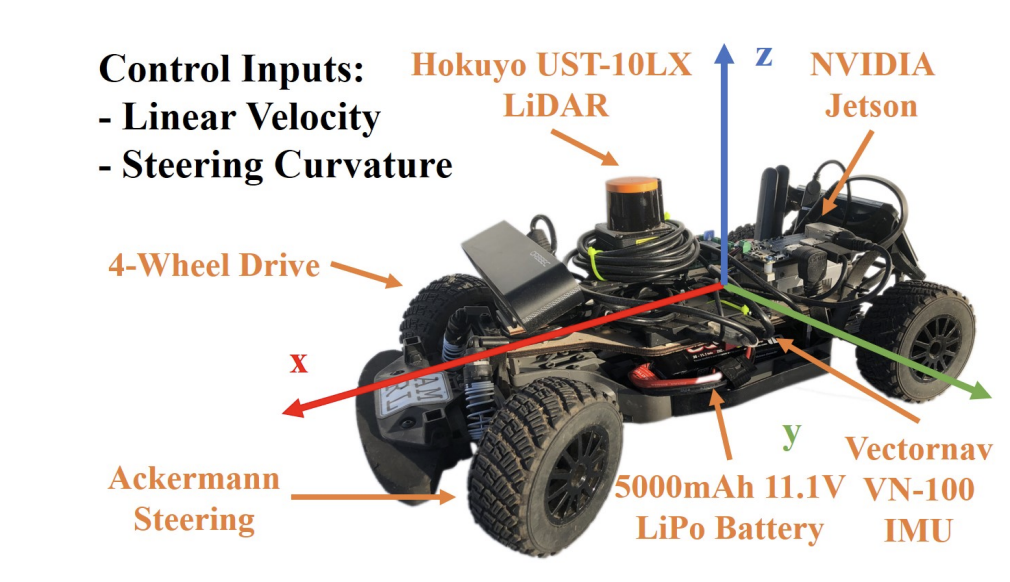}
  \caption{UT AUTOmata F1/10 scale vehicle, used for testing}
  \label{fig:car}
\end{figure}

We will first talk about the small-scale autonomous vehicle used to perform our testing. Shown below (Fig. \ref{fig:car}), we can see a detailed diagram of this vehicle, outfitted with a LiDAR sensor, an NVIDIA Jetson TX2, and a Vectornav VN-100 IMU (image courtesy from \cite{ikdhighsped}).
\par Furthermore, these vehicles use an Ackerman steering system, are four-wheel drive, and have a TRAXXAS Titan 550 motor controlled by a Flipsky VESC 4.12 50A motor controller. Onboard each F1/10 vehicle is copy of ROS (Robot Operating System), which is used to control and launch nodes for operating all core functionalities within the car (see \ref{ros:overview}).
\begin{figure}[h!]
  \centering
  \includegraphics[width=0.42\textwidth]{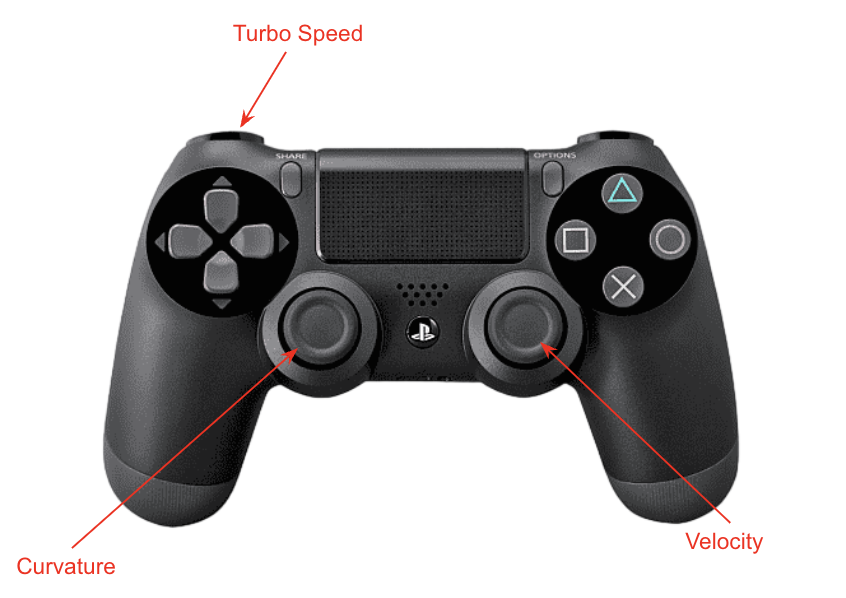}
  \caption{{PS4 Controller, used for teleoperating the vehicle}}
  \label{fig:ps4}
\end{figure}
\par Shown in Fig. \ref{fig:ps4} is the PS4 controller we used to teleoperate this vehicle. As part of the car's software, it is required for manual mode, and when operating the car with autonomous commands, it is required to hold down a button that can be released if the code needs to instantly be stopped (essentially a failsafe).

\subsection{Software}
We would also like to give a bit of overview on the software that we are running in this car.
\subsubsection{Overview of Robot Operating System (ROS)}\label{ros:overview}
Onboard each of these F1/10 UT AUTOmata vehicles is a copy of ROS, loaded onto Ubuntu 18.04. As a little bit of background, we woud like to discuss ROS at a high level. If you are already familiar with ROS, please feel free to skip over this section.
\begin{enumerate}
\item[(a)] ROS provides libraries and tools to help software developers create robot applications. Messages are routed via a transport system with publish/subscribe semantics.
\item[(b)] A \textit{node} (commonly referred to as \verb|rosnode|)  sends out a message by publishing it to a given topic; the \textit{topic} is a name that is used to identify the content of the message. This \textit{node} that is interested in a certain kind of data will subscribe to the appropriate topic. There may be multiple concurrent publishers and subscribers for a single topic, and a single node may publish and/or subscribe to multiple topics. 
\item[(c)] In general, publishers and subscribers are not aware of each others existence. The idea is to decouple the production of information from its consumption. Logically, one can think of a topic as a strongly typed message bus; each bus has a name, and anyone can connect to the bus to send or receive messages as long as they are the right type.
\end{enumerate}
\subsubsection{Vectornav IMU}
The IMU onboard each of these cars is a VN-100 IMU, which is not standard compliable with ROS. For this, we had to use a special script specified by UT AMRL for launching a \verb|rosnode| that corresponds to the Vectornav, allowing us to poll data off the onboard computer. Note that when booting the car, there are launch scripts that start the car's relevant nodes (\verb|gui|, \verb|vesc_driver|, etc.), but the Vectornav node is never launched. Thus, it must be done manually by downloading the Vectornav VN-100 scripts and launching it through \verb|roslaunch vectornav.launch|.
\subsubsection{Testbed for Navigation}
For our use, we duplicated and re-used a fresh copy of the CS378 starter code (provided copy for reference at \cite{amrlgit}). We forked a copy of this over to our own GitHub, and built on top of the existing \verb|navigation.cc| file in this repository. At a high level, commands are executed at a frequency of \textit{20 Hz} on the vehicle in the navigation \verb|src| within the main \verb|Run()| loop; a singular command is repeated 20 times per second to be executed on the actual car, by publishing a message to the drive node. At test time, we take some input velocity and curvature ($v_i$, $c_i$), convert the curvature ($c_i$) into angular velocity ($av_i$), and then feed ($v_i$, $av_i$) into our neural network. The corrected angular velocity is predicted by our neural network, and gives an output of $av'$, which we convert back into curvature ($c'$) and feed as a control input onto the onboard computer. This is discussed in depth in section \ref{nn}.
\subsubsection{Libtorch Model Loading} Once we collect our data, which is discussed in sections \ref{data} and \ref{datacollect}, we offload all the \verb|.bag| files to our local laptops for data training. We utilize the PyTorch library and Python 3.10.8 on M1 Pro silicon chips (2021 MacBook Pro 14-inch, 16-inch). Training will also be discussed more in depth in section \ref{training}. As the codebase that we are working with is in C++, the model that we trained from PyTorch (\verb|saved_model.pt|) had to exported to a libtorch model so that we could load it in on our car. We took the tracing code from \cite{ikdhighsped} and turned it into a script that loads in the model, performs a trace, and then saves it. This traced model was then transferred over using \verb|scp| to the actual onboard computer, loaded in the relevant files, and then passed the right input tensors. The dimension set within the loading is ${1, 2}$, representing a 1x2 vector, exactly similar to what we feed in at training time. We noticed that the libtorch model, once loaded, was efficient and quick at providing reliable outputs.

\subsection{Data Extraction} \label{data}


Data is inherently one of the most important parts to training our IKD model, and as such, must be cleaned and pre-processed before being fed into the model. To collect data, one can perform a command known as \verb|rosbag record|, in which you specify the topics to record. In our case, our model was only being trained off the joystick and the inertial IMU data, so the only two topics that we needed to record (or subscribe to) are \verb|\joystick| and \verb|\vectornav\IMU|. This records a \verb|bagfile| that efficiently records the required topics, which will later be converted to a CSV for data loading. In another sense, we are recording the joystick linear velocity, the joystick angular velocity, and the actual IMU angular velocity in the \textit{z} direction. 
\subsubsection{Conversion of Bag Files to CSV Format}
It is important to note that the data was also not clean in its original format. This is understandable; we had to convert the bag file to a CSV (script accessible in \cite{autogit}) so that it could be processed properly as readable data. This conversion simply extracted the bag file data into a CSV, leaving us with two files: \verb|slash_joystick.csv| and \verb|slash_IMU.csv|. Some bag files had more than 6-7 minutes (35,000+ data points) of recorded data.
\subsubsection{Cleaning CSV Files and Aligning Values}
The next step of looking at these two CSV files was to clean and align them. The first step of cleaning involved trimming any values at the beginning or the end of the dataset; the recordings for bag files are at \textit{40 Hz}, which means that from the time that the \verb|rosbag record| command is executed on our laptop and we actually start teleoperating the car, there is a bit of a delay between the recording. Thus, we want to trim out these values that are zeros at the beginning and end of the CSV files.
\par After this, there is the problem of actually aligning these files. This is also a very important part; we noticed that at different velocities, there are different IMU delays. To account for this, we had to analyze both CSV files and find the actual IMU delay value (e.g. at \textit{1.0 m/s}, the optimal IMU delay was .176 for our data). We leverage an alignment script that extracts the IMU measurements and joystick data from each CSV file, computes the delay between the two sets of data by minimizing the error between their velocities, and finds the best matching IMU angular velocities by traversing through potential delay values. 
\par An alternative way of doing this is that we can create a plot to visualize the error between the delay values, and choose the lowest off the graph. Alternatively, the alignment function was very useful in helping us find corrupted bag files, as the CSV files would generate very bizarre "optimal" IMU delay values such as -.84 vs. the normal range they are expected to be in (.17 - .21).
\par Note that linear velocity was just being measured off the joystick from the PS4 controller, thus it needed no alignment with the IMU, just pairwise alignment with the joystick angular velocity.
\subsubsection{Writing Train Data} Once we obtained all the optimal imu-delay values, we created separate training data files for every single CSV. We create an array of evenly spaced time points, used interpolation to estimate the joystick velocities, rotational velocities and IMU angular velocities, and appended the data to their respective lists for a dump to a training data file.
\subsubsection{Merging Training Data} 
\begin{figure}[h!]
  \centering
  \includegraphics[width=0.42\textwidth]{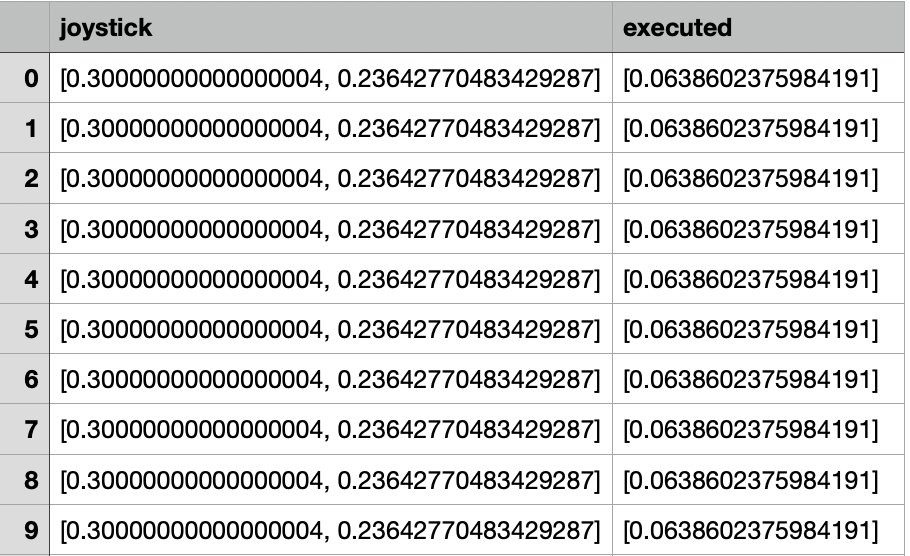}
  \caption{Example sample from our training data}
  \label{fig:example-of-training-data}
\end{figure}

In the case in which merging data is necessary for training our model, we have created a custom script that reads in all training data files, concatenates them into a single data-set, adjusts the indexing and saves to a fresh file. Refer to the figure to see an example sample from of our training data set \ref{fig:example-of-training-data}.

\subsection{Data Collection} \label{datacollect}

In this section, we outline all the categories of data we collected, graphical insights into them, and how they helped us during our work. 

\subsubsection{Turning Data}

\begin{figure}[h!]
  \centering
  \includegraphics[width=0.42\textwidth]{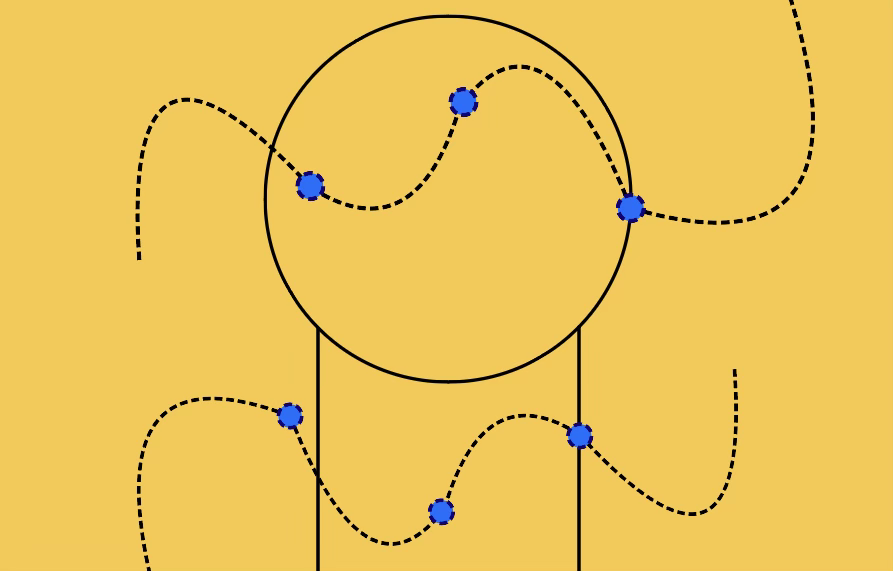}
  \caption{Diagram of the Anna Hist Gymnasium with example turning trajectories of the vehicle during the 5-min timers.}
  \label{fig:turning-diag}
\end{figure}

For our first category of data collection, we obtained an extensive range of turning data. We made the decision to collect 5 minutes of data for every speed (1.0 through 5.0), for a total of 5 minutes. We made the decision to exclude 6.0 and 7.0 from collecting purely sharp turning data, as the dimensional bounds of our research space and car speed made it extremely difficult to do simple sharp turns without heavy slippage or risk of hitting an object within the room. For turning data, since we were limited to the bounds of the room, there were only 3-4 turning points before we reached the ending wall of the room. The graphic displays the general pattern of the vehicle movement during data collection \ref{fig:turning-diag}.

\par After we completed this data collection, we wanted to ensure that the training data was following the expected behavior that we were seeking to correct with our inverse kinodynamic learning. Let us show the charts first, and then explain the context behind them.

\begin{center}
\begin{figure}[h!]
  \centering
\includegraphics[width=.42\textwidth]{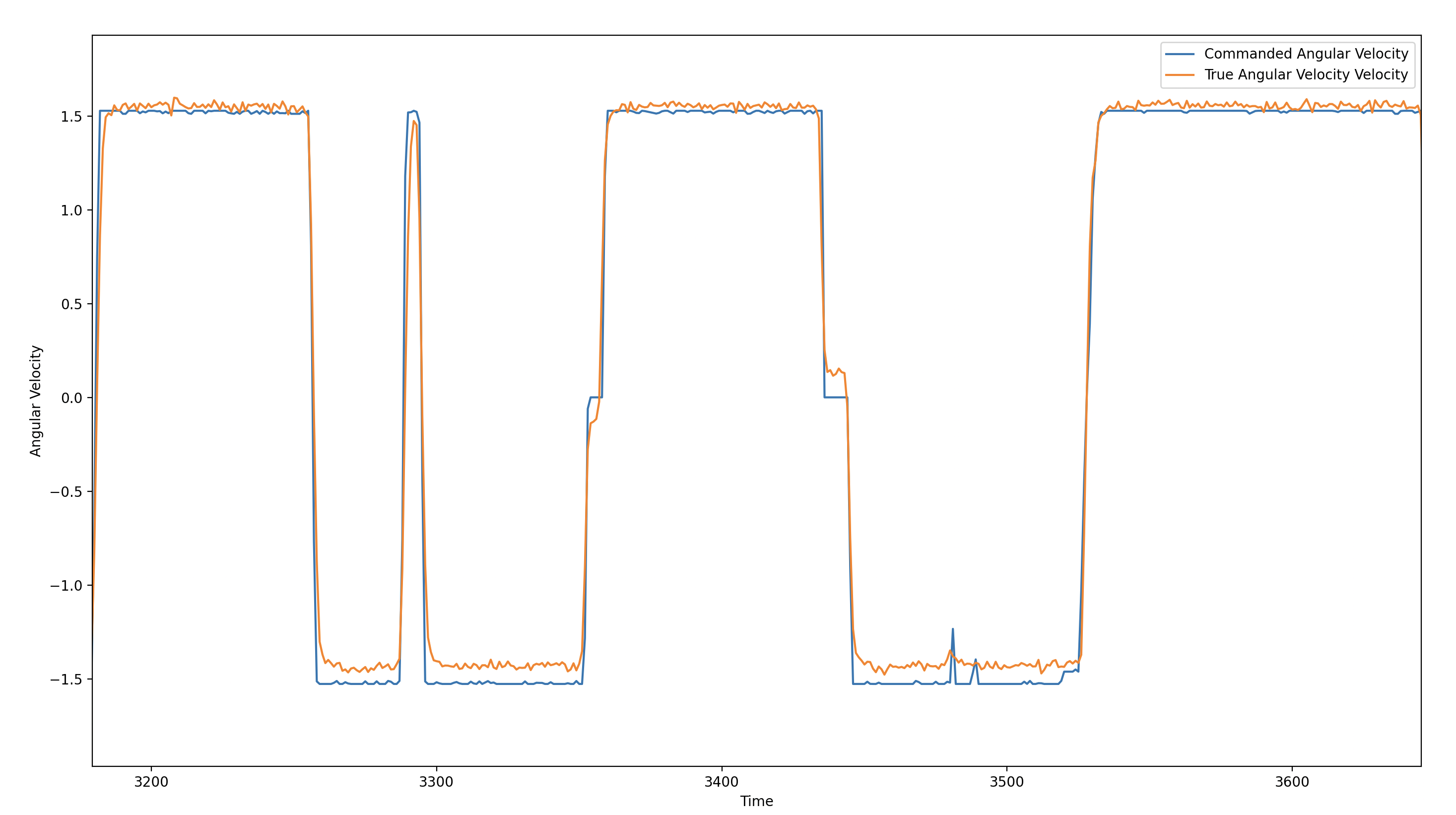}
\caption {Data with 2.0 m/s, Sample of Graph plotting AV and CAV with respect to T.}
  \label{fig: 2.0 turning av}
\end{figure}
\end{center}

\par
\begin{center}
\begin{figure}[h!]
  \centering
\includegraphics[width=.42\textwidth]{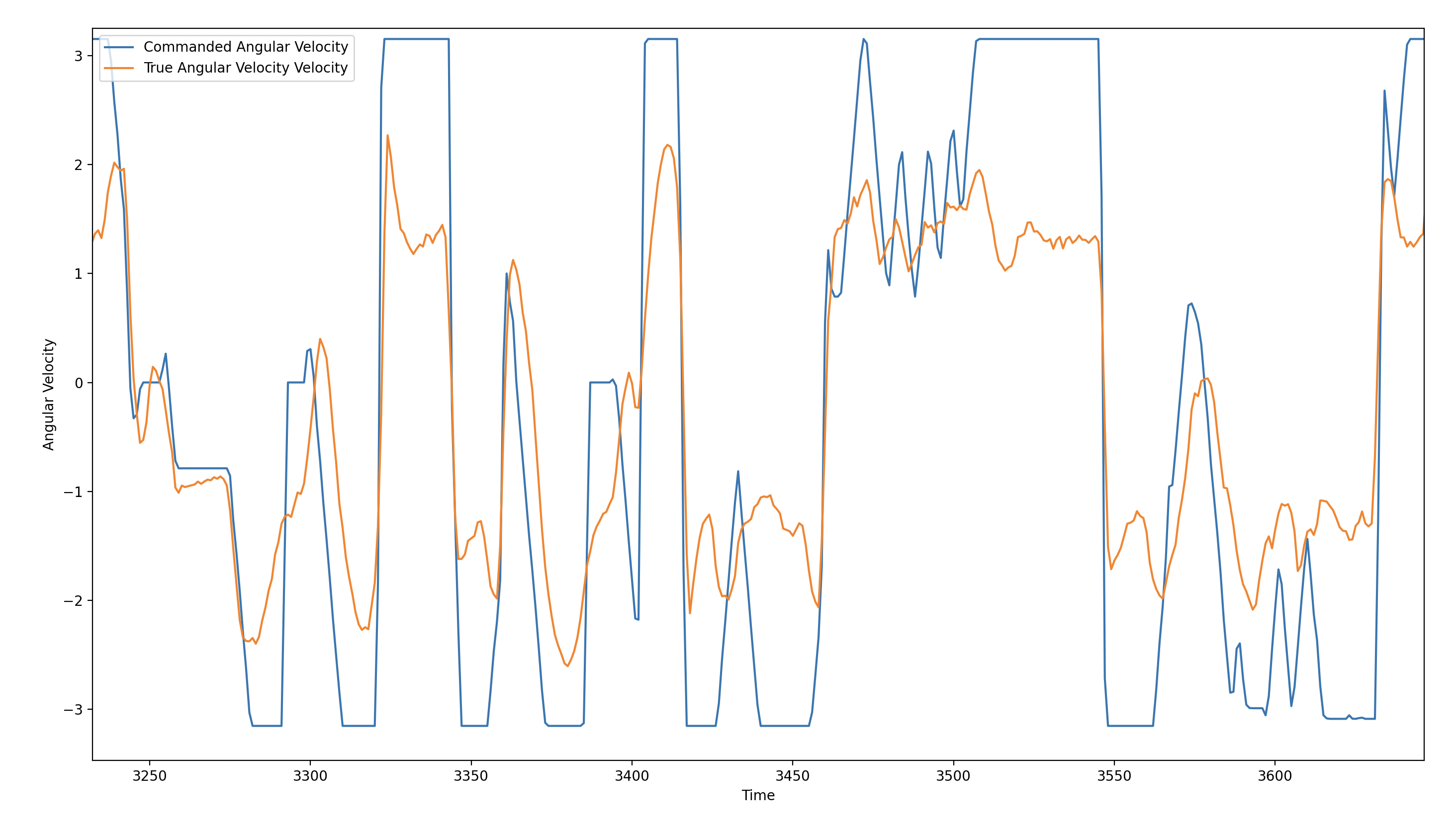}
\caption {Data with 4.0 m/s, Sample of Graph plotting AV and CAV with respect to T.}
  \label{fig: 4.0 turning av}
\end{figure}
\end{center}

 In the chart above \ref{fig: 2.0 turning av}, we have plotted the true angular velocity and the commanded angular velocity with respect to time. What is being shown here is a snippet of time for clarity. The orange line represents the true angular velocity from the inertial measurement, and the blue line represents the commanded angular velocity from the joystick. At low speeds, we can see that the difference between the two is extremely minimal. This is quite expected behavior. But, let us now look at a graph with turning data at 4.0 m/s \ref{fig: 4.0 turning av}.

\par
\begin{center}
\begin{figure}[h!]
  \centering
\includegraphics[width=.42\textwidth]{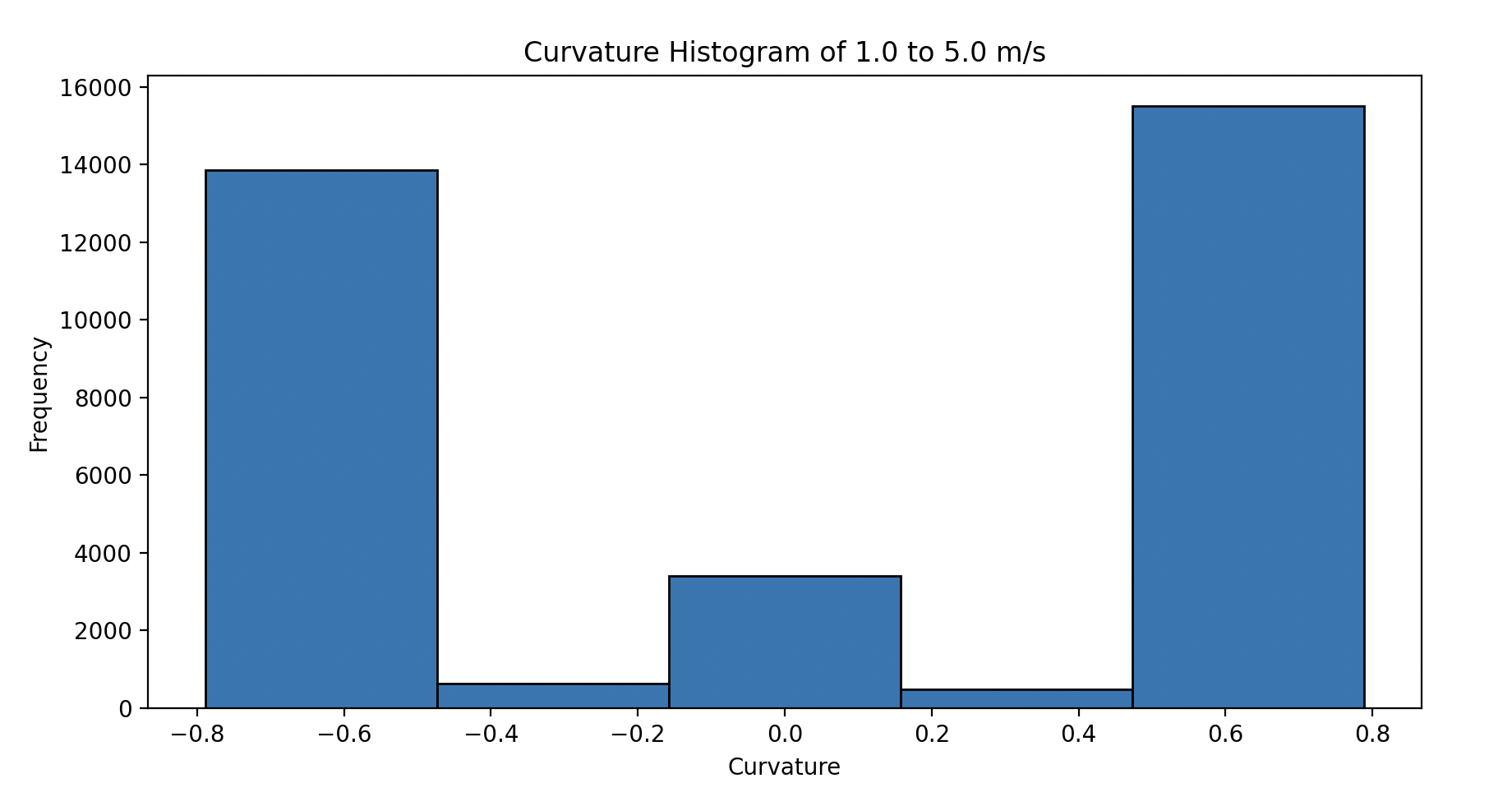}
\caption {Histogram with curvatures binned. Tail-ends are noticeably skewed, with minimal frequency in middle curvatures.}
  \label{fig: merged-curvature-turning}
\end{figure}
\end{center}

 Upon examining this alternative representation, it is very evident that there is a difference between the IMU angular velocity and the joystick commanded velocity. As speed increases, the divergence between the two becomes more glaringly obvious. This visualization technique from ground-zero proved to be very useful, and helped us find several inconsistencies in our data-set and catch problems with training. We also have all alternative charts such as (3.0, 5.0, 1.0) in a directory on our GitHub. Furthermore, before collecting more data, we sought out to examine the turning data linear velocities and curvature, as these are essential parameters in the creation and execution of the autonomous navigational script. We made this decision after attempting to make corrections with the trained IKD model and noticed values that were extremely off, especially for random inputs. We first merged all the turning data files together using a similar structure to our merging script mentioned in the data extraction section, calculated curvatures, and binned them together in a histogram style format \ref{fig: merged-curvature-turning}. Curvatures were calculated using the simple mathematical formulation that can be found in the subsequent section.

\par
\begin{center}
\begin{figure}[h!]
  \centering
\includegraphics[width=.42\textwidth]{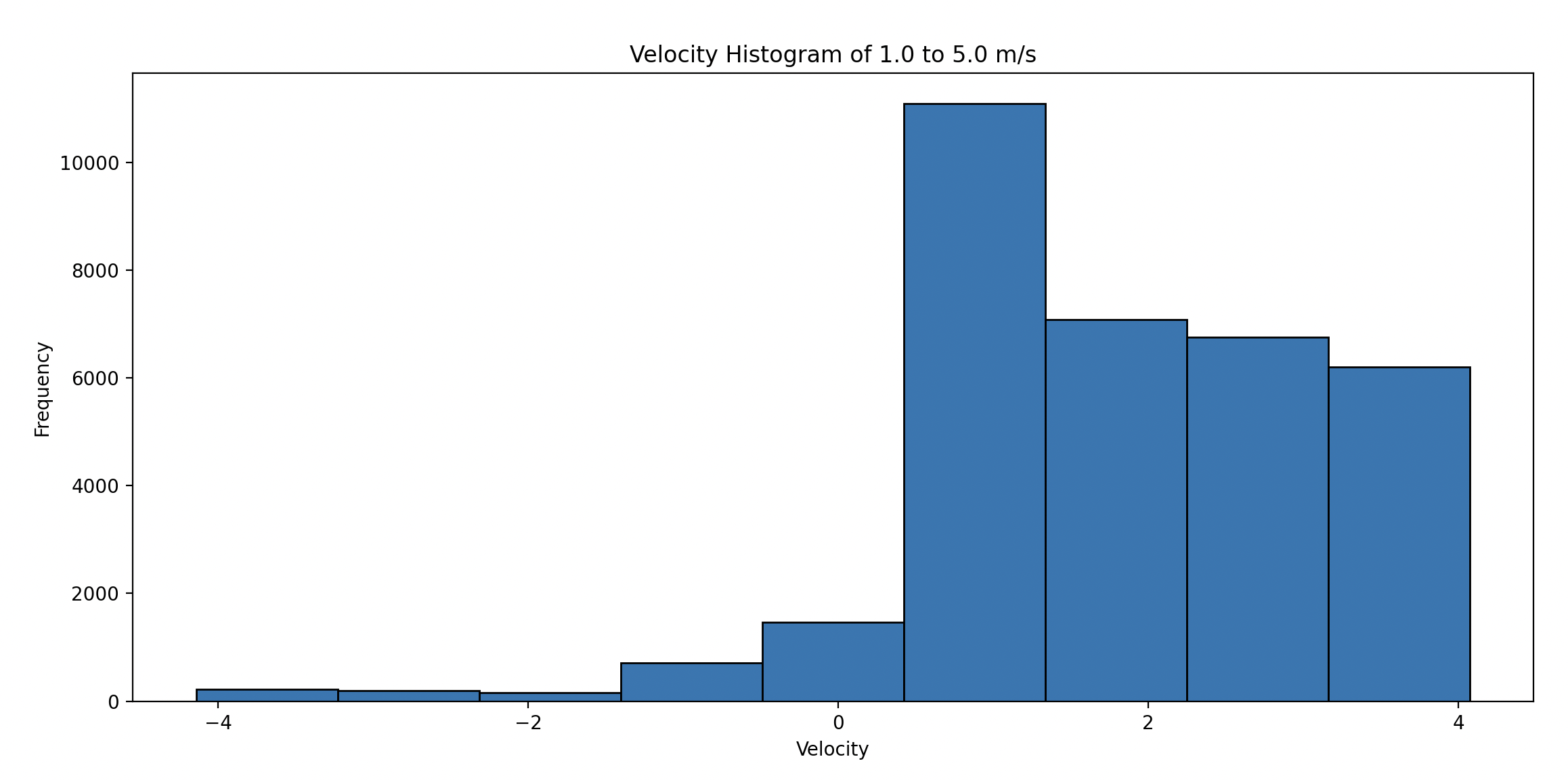}
\caption {Velocities binned on an intentional extensive range. Represents 1.0 to 5.0 m/s turning data.}
  \label{fig: merged-velocity-hist}
\end{figure}
\end{center}

 By just a visual analysis, we were quite surprised by the resulting histogram. We found that when joy-sticking the UT AUTOmata car, we could only really turn at max curvatures, as turning with middle curvatures is difficult on the joystick, and quite impractical. Being that our overarching task is to reduce the slip on a drift, we are generally more focused on collecting data for tighter turns and curvatures anyways, but, for the sake of our circle experimentation, we did want to potentially diversify the data-set to contain more middle valued curvatures. During this phase of data collection, we also wanted to get a more precise insight into the patterns of the velocities. We took the merged data used from the previous step, doubled the number of bins in comparison to the max value of velocity, and created a histogram \ref{fig: merged-velocity-hist}. The results were extremely fascinating, and opened a new door for data exploration.

\par
\begin{center}
\begin{figure}[h!]
  \centering
\includegraphics[width=.42\textwidth]{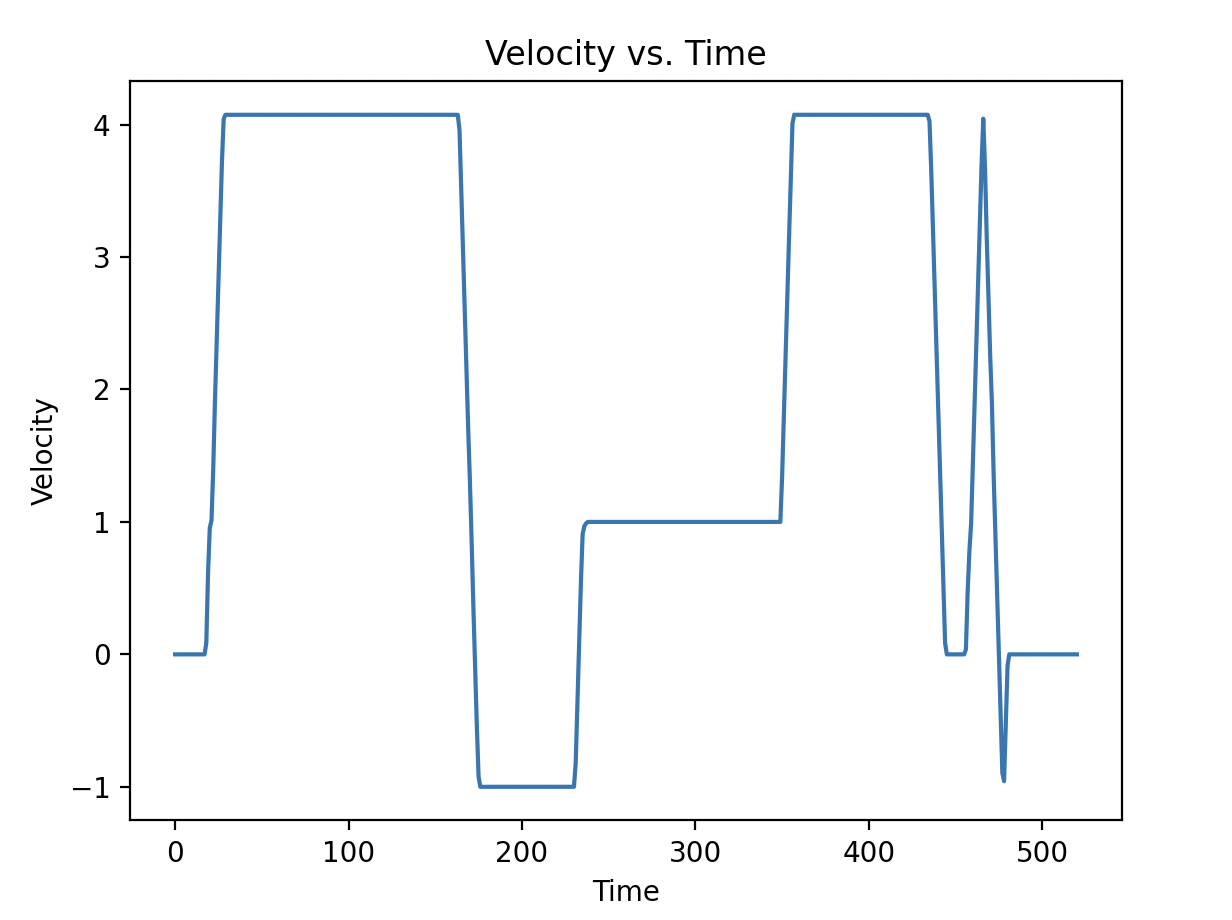}
\caption {Two turbo attempts on a long smooth sidewalk, with max speed set to 6.0.}
  \label{fig: joystick-turbo-exp}
\end{figure}
\end{center}

 We noticed that the velocities were capped at the low 4.0 range. We wanted to explore why the max linear velocity that was being extracted from our training data was 4.219, despite the turbo-speed during training data collection being 5.0. Our initial insight was that the car does not have enough space within the dimensions of the gymnasium to reach the maximum speed. To test this, we setup a small experiment outside on a long smooth sidewalk that was almost 4x the length of our research room. Within the vesc-driver configuration, we intentionally set the turbo speed of the vehicle to 6.0, to ensure that there would be a guarantee for the vehicle to exceed the 4.219 max value as long as it had enough space to accelerate and decelerate. The results can be seen in this figure \ref{fig: joystick-turbo-exp}.

\par It was evident that somehow, the velocity was being capped at 4.219 m/s. We knew that it was not the acceleration limit, as inside of our vesc-driver configuration, this was set to 6.0. But, we did realize that during the extraction of our joystick velocity, we leverage the ERPM gain and ERPM offset to clip the speeds in the training data. In the configuration, speed to erpm gain is set to a value of 5,356, which suggests to us that the motor has 2 poles and that the value was determined based on the gear configuration of the vehicles motor. Based on this logic alone, we knew that this is value that we should not touch in regards to the vehicle, as it tells the UT AUTOmata how many ticks of the motor correspond to certain distances, and could potentially ruin the car if changed. This gave us insight that the linear velocity was capped, which was not of big concern, as our trial runs prior to data collection for drifting proved that 4.2 linear velocity was plenty in obtaining an effective slippage on the gymnasium floor.

\subsubsection{Drifting Preface}

\par Before we discuss the collection of our drifting data, it is important to discuss the key tele-operation controls necessary to perform the drifting. Before we began recording rosbag files for drifting, we practiced the drifting on several different surfaces using our controller. The core idea behind drifting is that when a vehicle comes to turn (for example around an object), the weight of the vehicle shifts to the side that the car is turning, creating a sideways acceleration. In that moment, if done right, the tires grip the surface, and with enough speed, the acceleration can exceed the grip on the ground, leading to a continuous slip among the turn before regaining control of the vehicle. In our case, we leverage the turbo on the joystick to accelerate the vehicle towards a turn. The moment we begin turning, we cut off the acceleration from the turbo speed button on the controller and jerk the forward velocity joystick in the opposite direction. This sequence, combined with turning the wheels at the same time, leads to an over-steer effect and effective slip, creating a drift. We spent a great deal of time at the beginning of our work learning how to effectively tele-operate, as it was an important component of our data collection and potential future success of the IKD control correction.

\subsubsection{Loose Trajectory Drifting Data}

\par In this sequence of data collection, we decided to stick with a 5.0 turbo velocity. As per our discussion of the previous data collection, since we found that the speed is capped at around 4.2, 5.0 would be the only viable value to set within the vesc-driver configuration of our vehicle.

\par
\begin{center}
\begin{figure}[h!]
  \centering
\includegraphics[width=.42\textwidth]{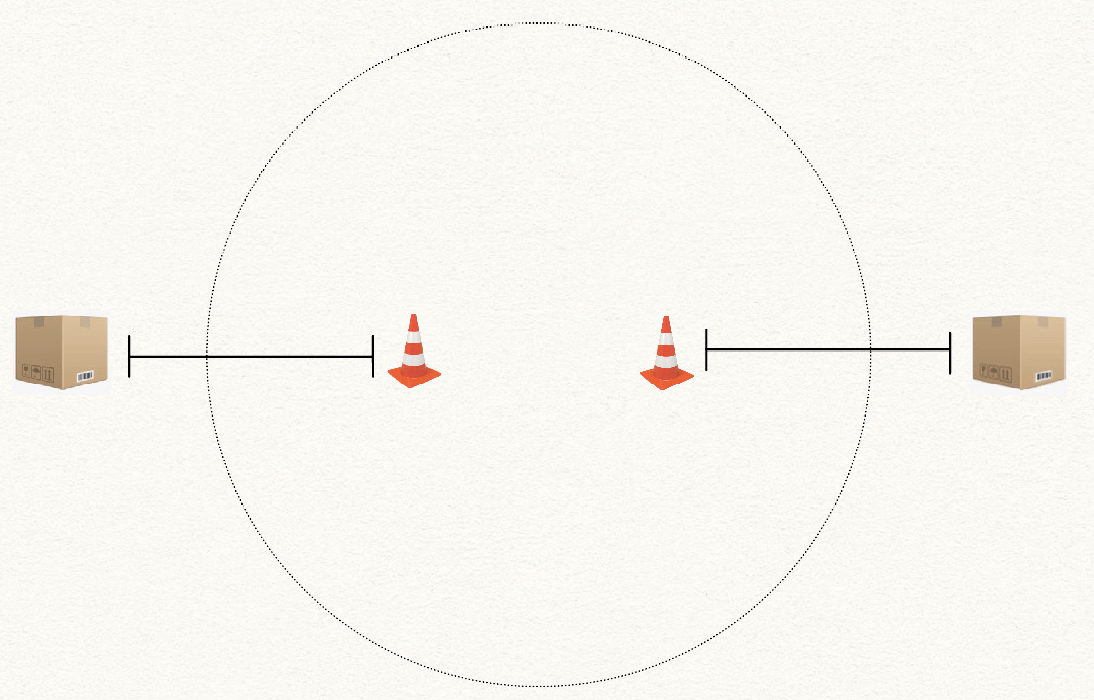}
\caption {Loose Drifting Setup: 2 cones and 2 boxes, each 2.13 meters apart. Turbo speed set to 5.0.}
  \label{fig: loose-drifting-diagram}
\end{figure}
\end{center}

 To collect data for this portion, we created a simple scenario to continually record drifting data. Please refer to the graphic. \ref{fig: loose-drifting-diagram}

\par
\begin{center}
\begin{figure}[h!]
  \centering
\includegraphics[width=.42\textwidth]{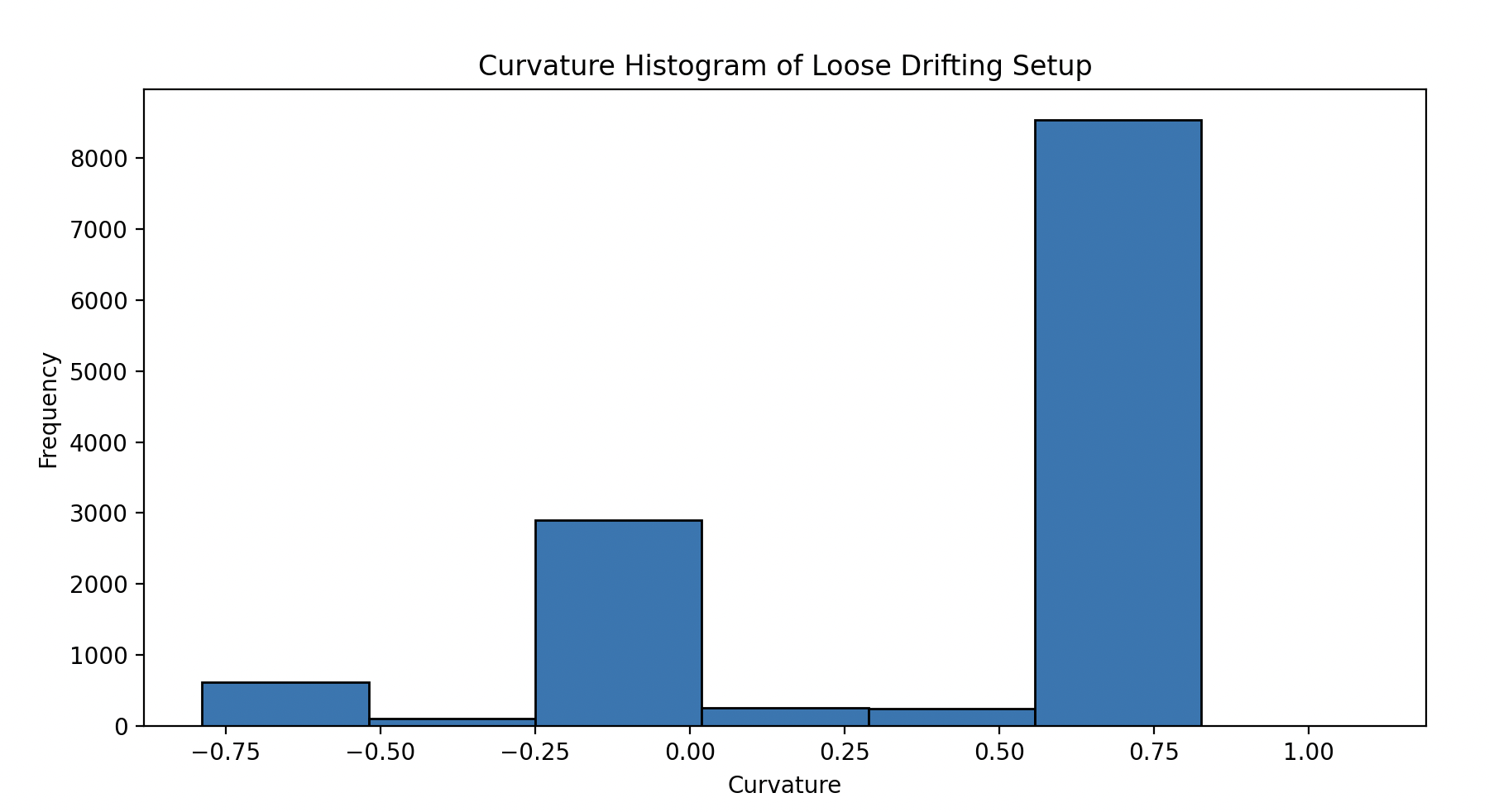}
\caption {Loose drifting data curvatures, approximately 10 minutes of data collection at a max 5.0 turbo speed.}
  \label{fig: loose-drifting-curvatures}
\end{figure}
\end{center}

 The setup consists of 2 AMRL testing cones, and 2 boxes that represent a "boundary" that the vehicle can not pass. The space between each box and cone on both the East and West sides of the graphic are 84 inches, or 2.13 meters. During teleoperation, the goal of the vehicle is to drift around the cone as tight as possible, but with low risk of impact due to the room for curvature and turning angle being so large. We thought of this as a way to diversify our dataset, since collecting extremely tight clearance drifts is complex, as we found in our second round of drifting experimentation. Additionally, this data collection consisted of 10 minutes straight tele-operation. As expected, we found that the curvatures were more oriented towards higher values, indicating that the tele-operation was decently robust \ref{fig: loose-drifting-curvatures}.

\par
\begin{center}
\begin{figure}[h!]
  \centering
\includegraphics[width=.42\textwidth]{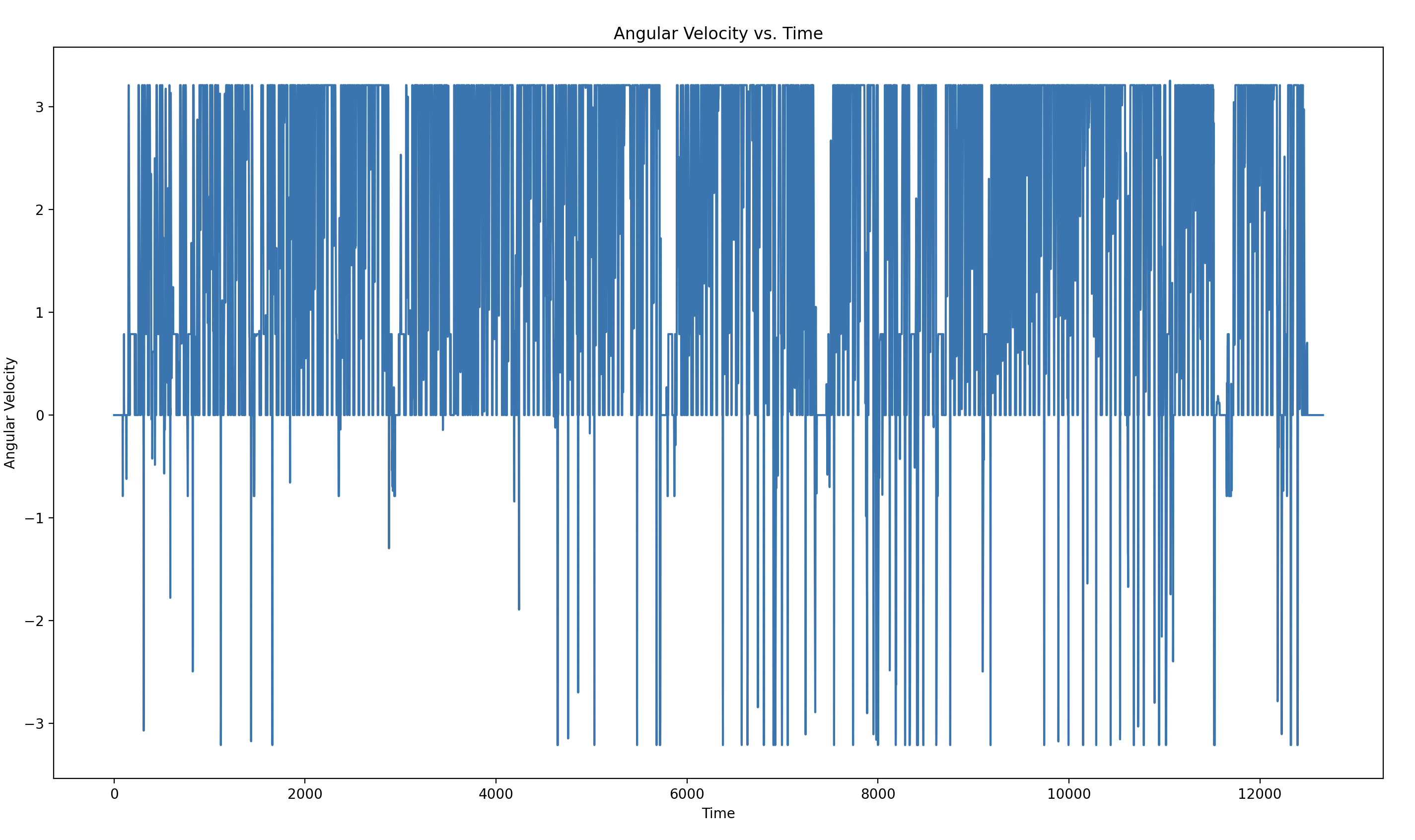}
\caption {Loose drifting data commanded angular velocity plotted against time.}
  \label{fig: loose-drifting-av}
\end{figure}
\end{center}

\begin{center}
\begin{figure}[h!]
  \centering
\includegraphics[width=.42\textwidth]{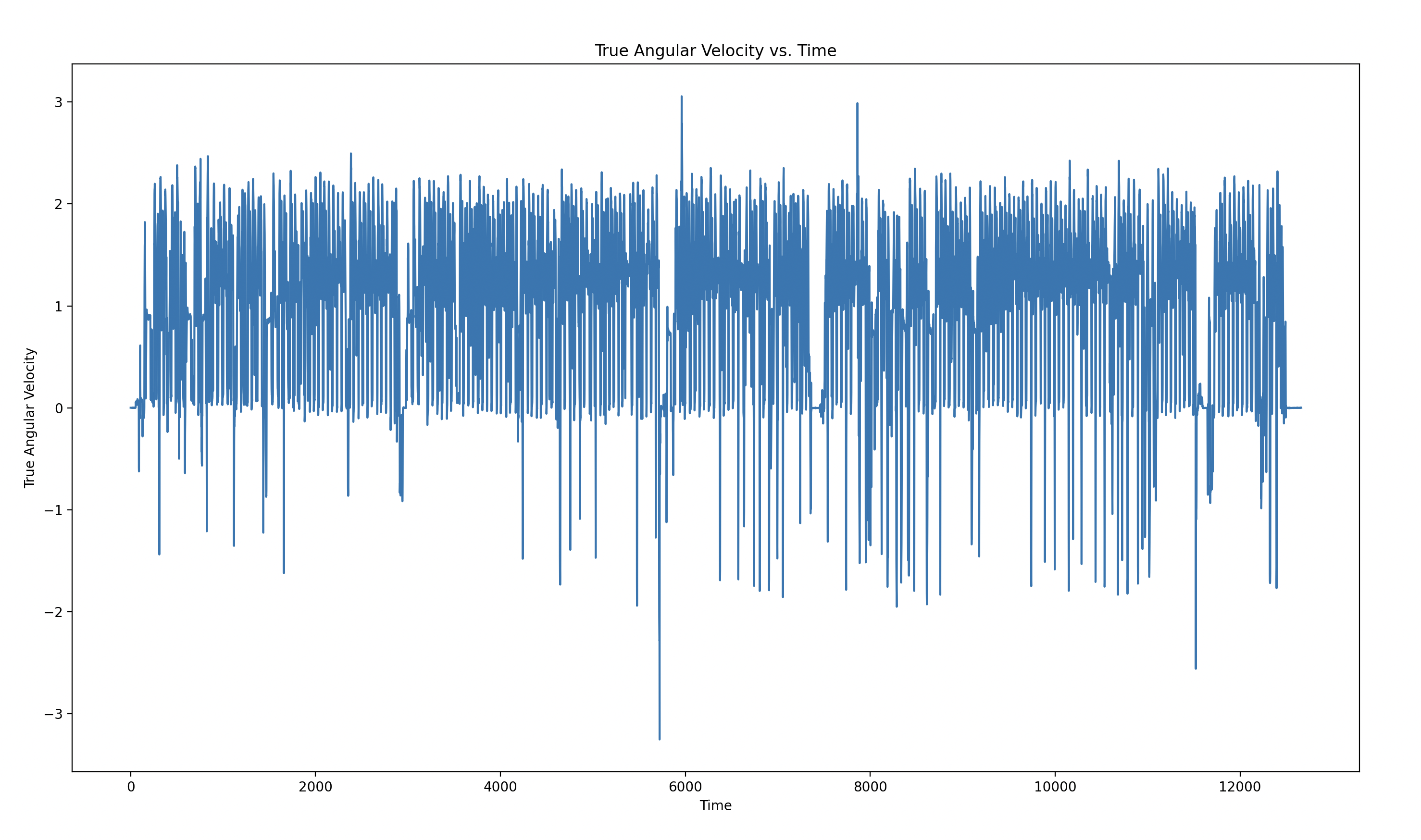}
\caption {Loose drifting data true angular velocity plotted against time.}
  \label{fig: loose-drifting-true-av}
\end{figure}
\end{center}

 Additionally, the angular velocities portrayed a very unique pattern when plotted. It seemed that the angular velocities recorded off the IMU had a much more clean pattern between the forward and backward drifts in comparison to the recorded commanded angular velocities. \ref{fig: loose-drifting-av} \ref{fig: loose-drifting-true-av}

\par
\begin{center}
\begin{figure}[h!]
  \centering
\includegraphics[width=.42\textwidth]{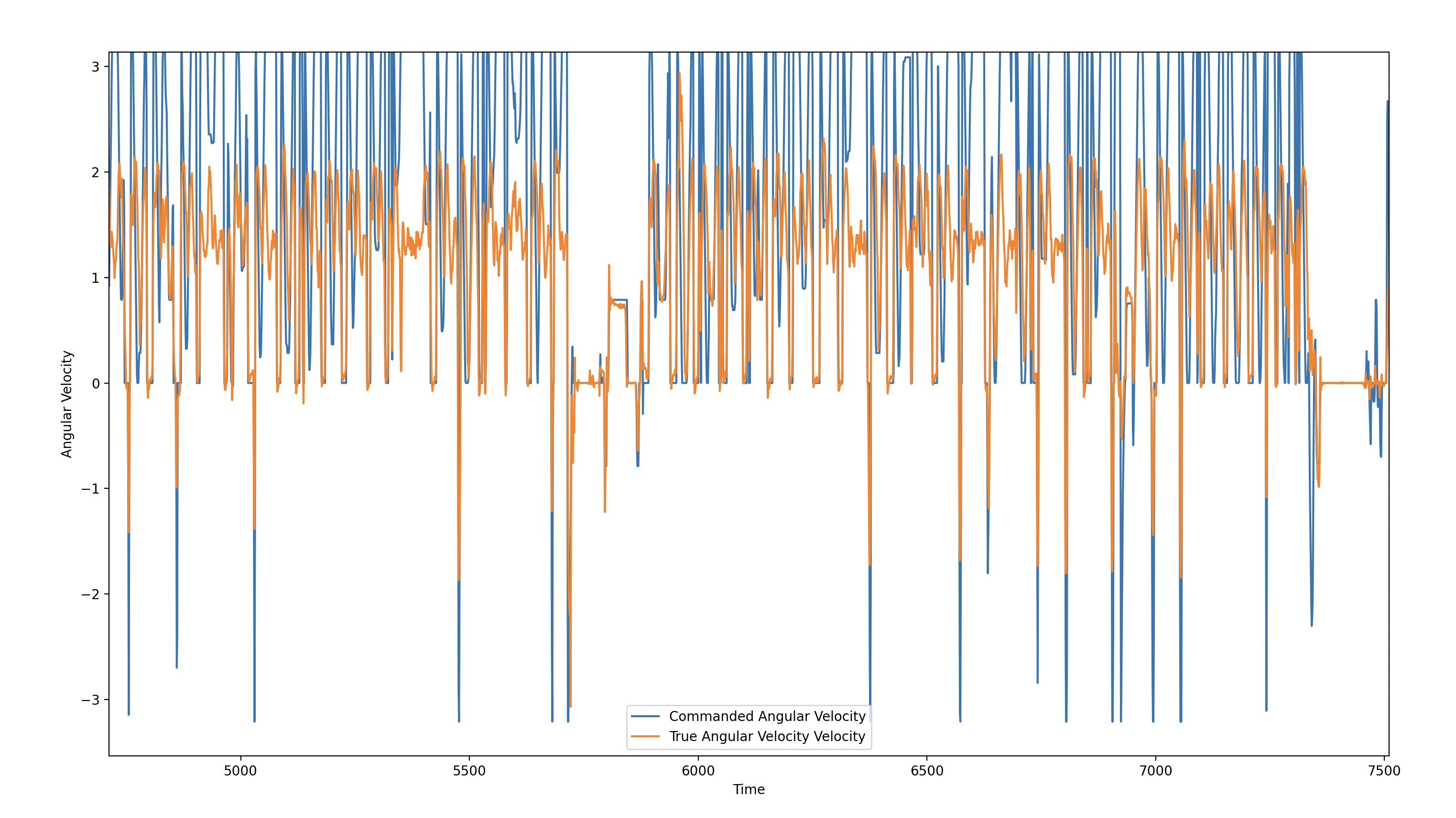}
\caption {Previous figures of angular velocities overlaid over one another.}
  \label{fig: loose-drifting-overlay-av}
\end{figure}
\end{center}

 And looking at the two overlayed on top of one another, it is very clear... there is a lot of room for angular velocity correction! \ref{fig: loose-drifting-overlay-av}

\subsubsection{Target Trajectory Drifting Data}

\par
\begin{center}
\begin{figure}[h!]
  \centering
\includegraphics[width=.42\textwidth]{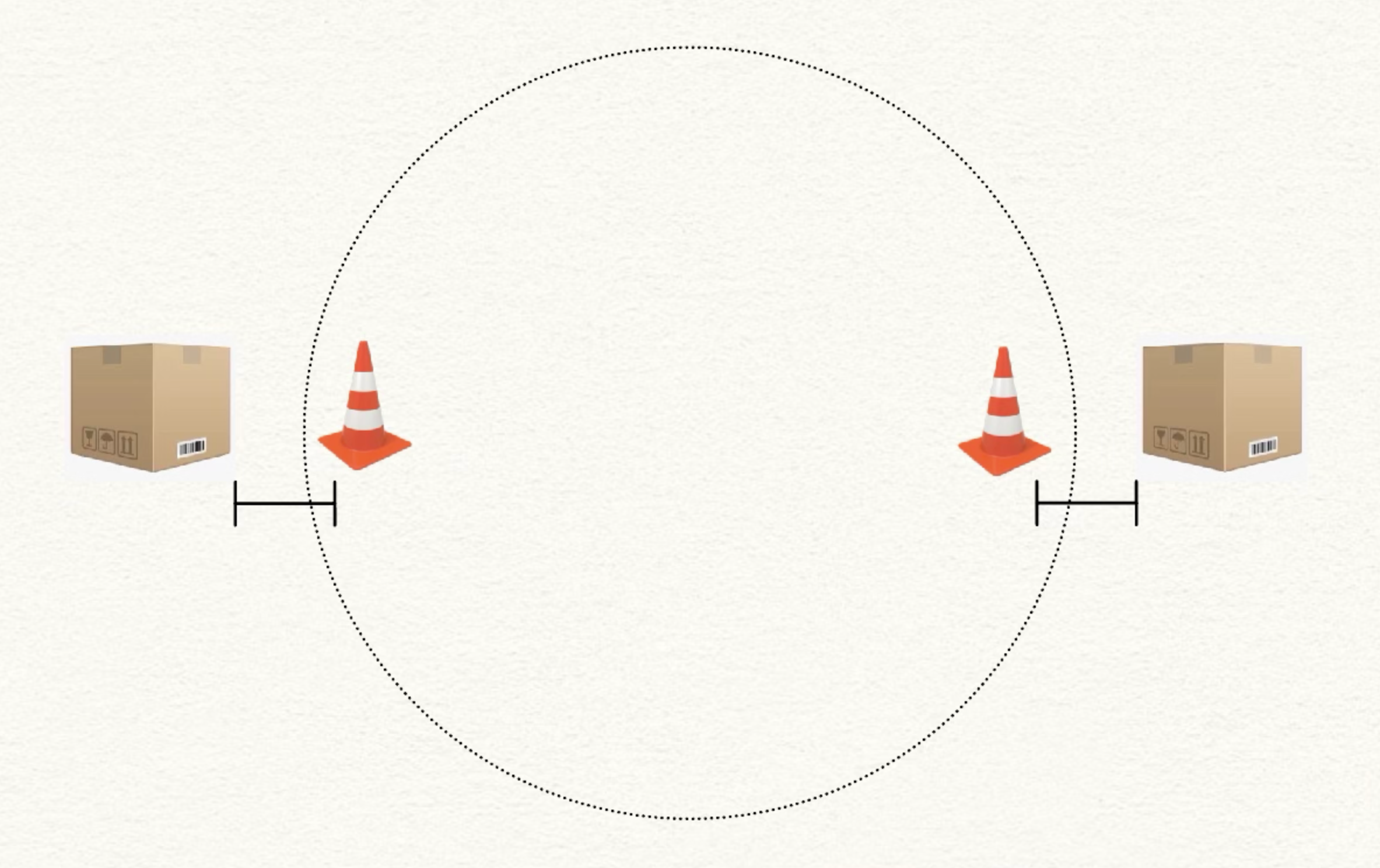}
\caption {Tight Drifting Setup (\#1): 2 cones and 2 boxes, each .81 meters apart. Turbo speed set to 5.0.}
  \label{fig: tight-simple-dirft}
\end{figure}
\end{center}

 In this sequence of data collection, we again decided to stick with a 5.0 turbo velocity. To collect data for this portion, we created two different scenarios. The first one was similar to the simple setup from the loose drifting data collection. Please refer to the graphic. \ref{fig: tight-simple-dirft}

\par
\begin{center}
\begin{figure}[h!]
  \centering
\includegraphics[width=.42\textwidth]{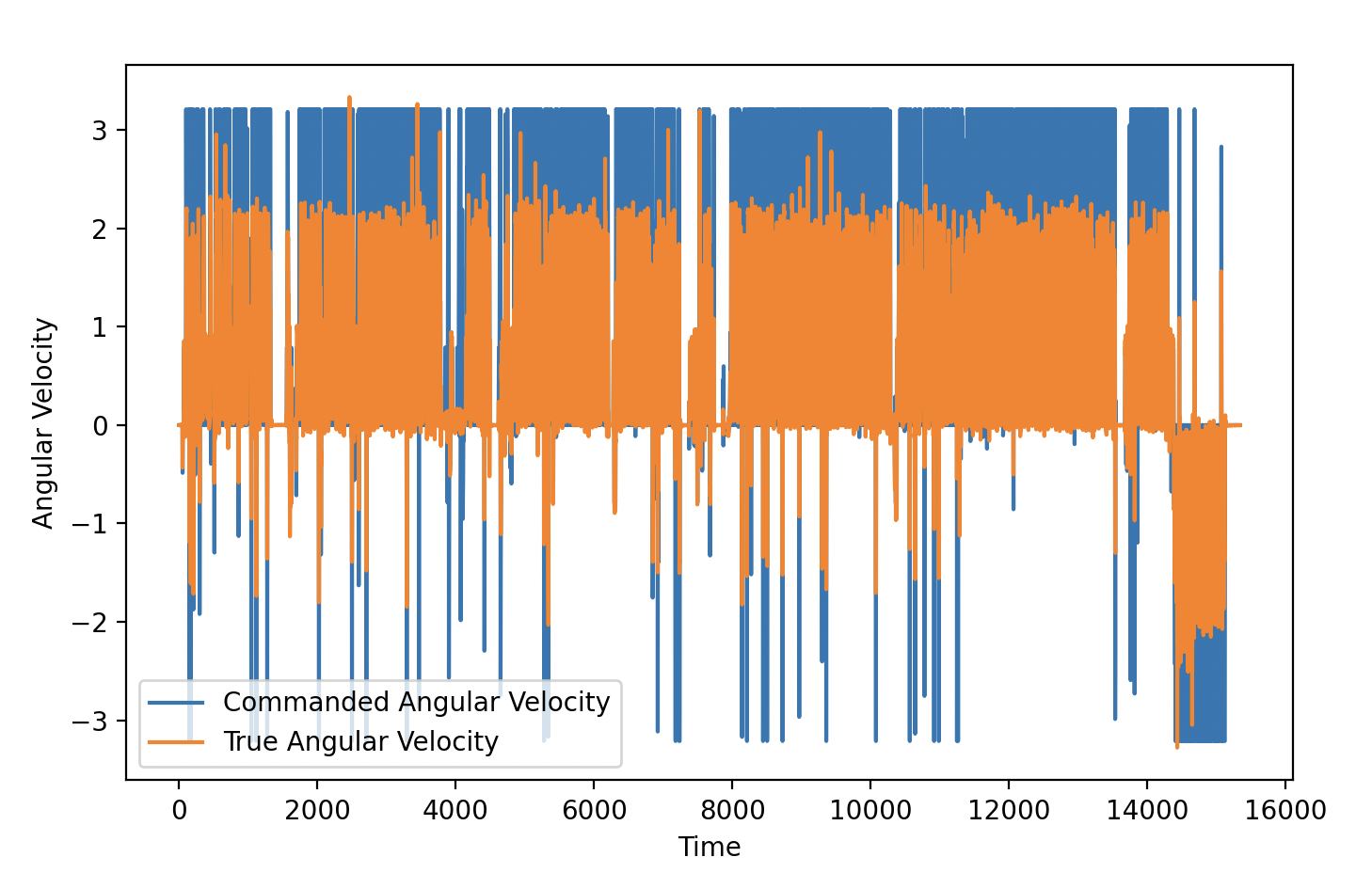}
\caption {Tight Drifting Setup (\#1): ALL angular velocities overlaid.}
  \label{fig: tight-av-overlay}
\end{figure}
\end{center}

 This setup (similar to the previous one), consisted of 2 AMRL testing cones, and 2 boxes that represented a boundary that the vehicle can not pass. For the spacing, we strategically chose 32 inches, or .81 meters for the distance between the cone and the box. The reasoning for this is that the car itself measures in at 19 inches, or .48 meters. We wanted there to be enough space for the car to be completely perpendicular to its original trajectory prior to entering the slip, and just a fractional bit of free space since teleoperation can not be perfect and we do not want to hit the box on every single instance we try to clear the turn around the cones. This data collection consisted of 10 minutes of tele-operation. Looking at the overlay between the commanded angular velocity and true angular velocity, it follows a very similar pattern.  \ref{fig: tight-av-overlay}.

\par
\begin{center}
\begin{figure}[h!]
  \centering
\includegraphics[width=.35\textwidth]{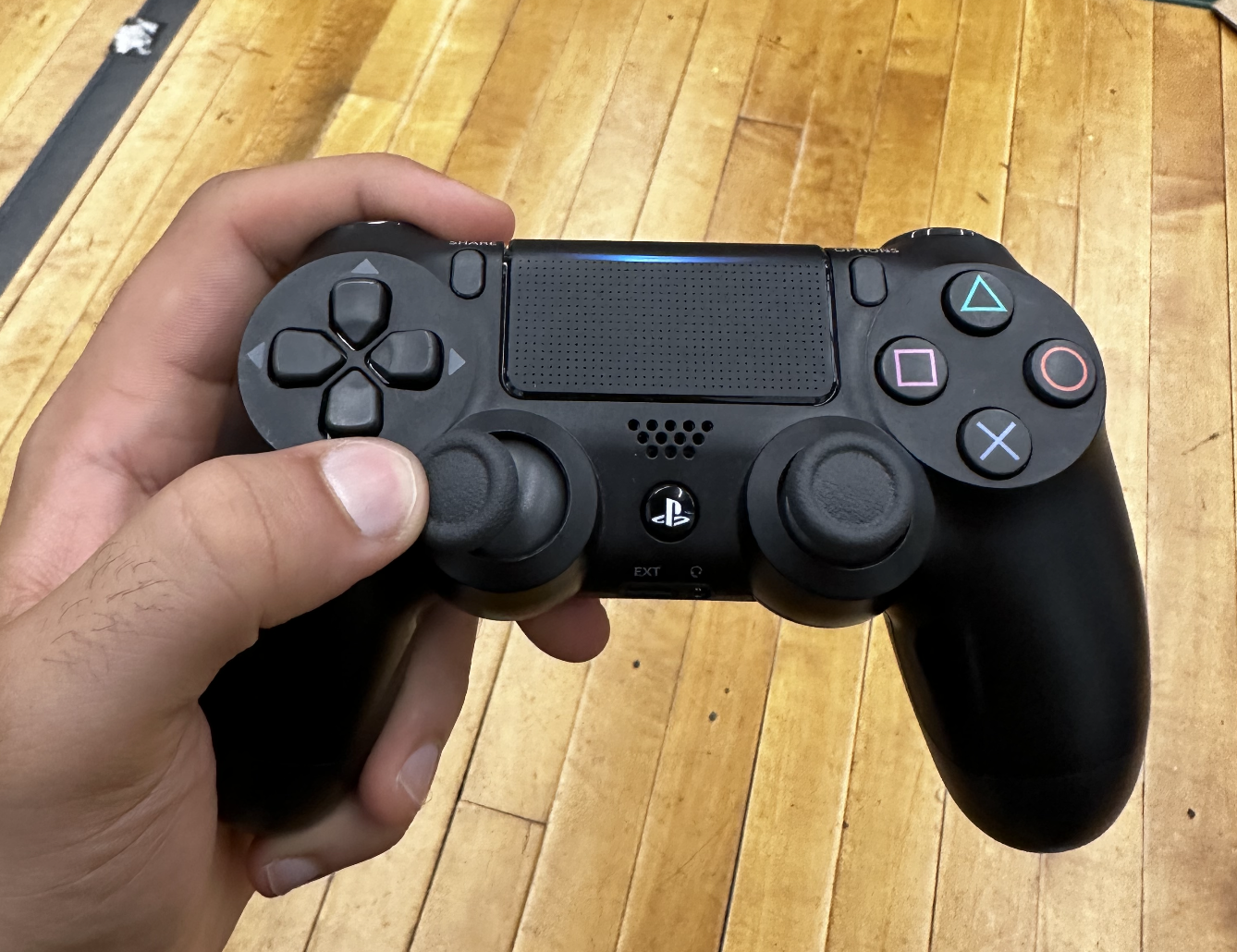}
\caption {PS4 Controller being held. Depicts the direction in which the bottom-left joystick is commanded during drifts.}
  \label{fig: drifting-commanding-example}
\end{figure}
\end{center}

 We would like to draw attention to an important observation regarding this graph. At the end of our data collection, we intentionally did something different from our normal data collection. Pay attention to time-step 14,000, and notice how our angular velocities are now majority in the negatives. \ref{fig: tight-av-overlay} Looking at all previous graphs, it is evident from most of these graphs that the angular velocities tend to be skewed towards the positive range. This is primarily due to the fact that the vehicle's rotation during our data collection occurred in a counter-clockwise direction. Conversely, when the rotation of the vehicle happens in a clockwise direction during slip or turn, the angular velocity becomes negative. During experimentation, we intentionally focus on the counter-clockwise direction, since that is the primary data our model is trained on. To provide a better understanding, we have included an image that depicts the direction in which the bottom-left joystick is commanded during drifts. \ref{fig: drifting-commanding-example}

\par
\begin{center}
\begin{figure}[h!]
  \centering
\includegraphics[width=.35\textwidth]{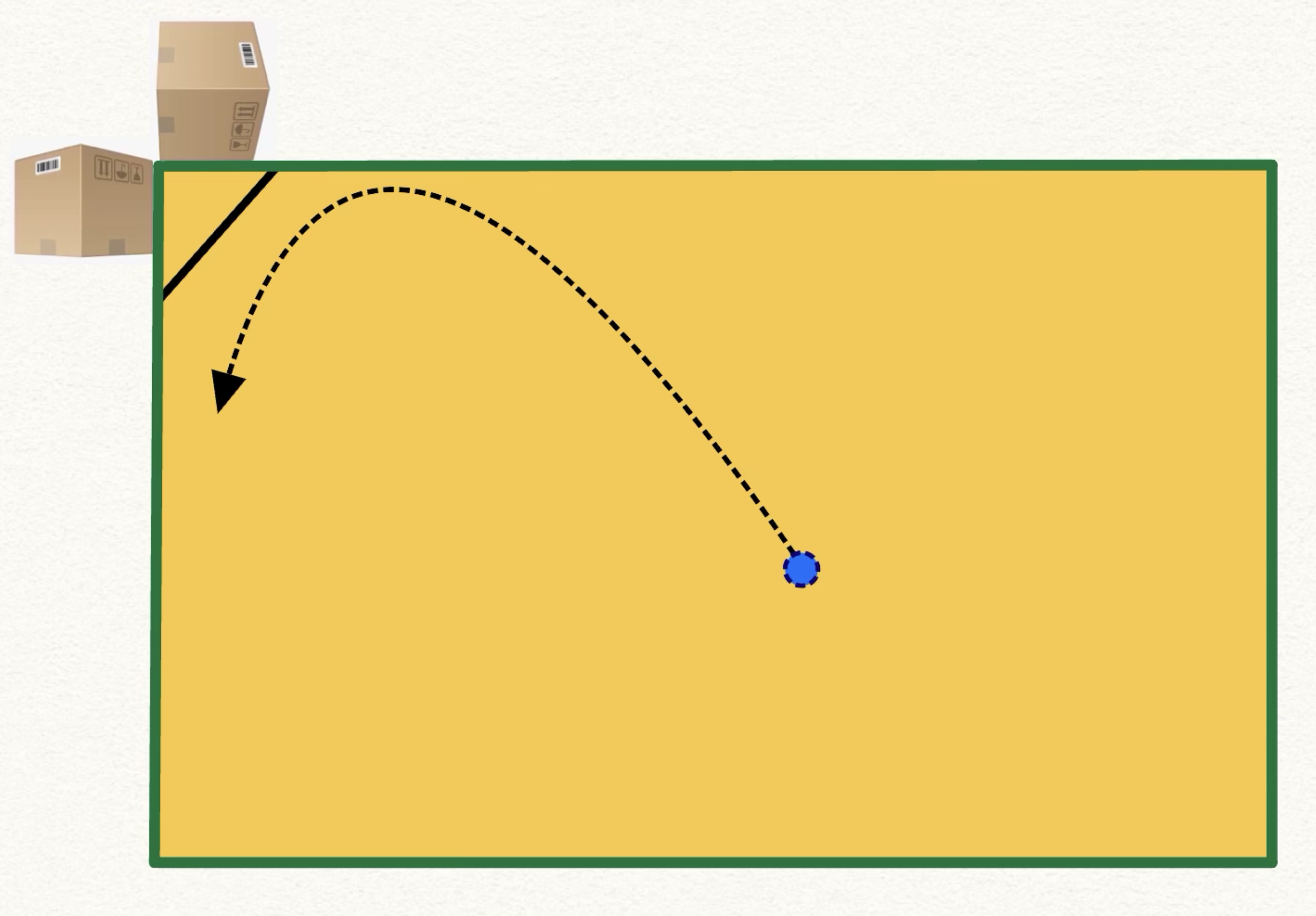}
\caption {Graphic of experimentation and tight data collection \#2 setup.}
  \label{fig: experiment-drifting-trajectory-graphic}
\end{figure}
\end{center}

 Overall, we found that drifting around the cone with such a tight space of clearance was difficult. To obtain more tight drifting data, we leveraged our experimentation setup to get more training data for our neural network. \ref{fig: experiment-drifting-trajectory-graphic}

\par In the graphic \ref{fig: experiment-drifting-trajectory-graphic}, the blue dot represents UT AUTOmata vehicle, the dotted line represents an example trajectory, with the black line being approximately the size of the vehicle. On a perfect slip, the right lateral side of the vehicle should align parallel to the black line at some time point through the drift, and then continue forward. This is an extremely tight curvature, and requires close to perfect execution to avoid hitting the boxes, especially when moving at a turbo speed of 5.0 meters per second. We collected a total of 15 minutes of data in relation to this scenario. We intentionally collected more data for this as the success rate of our tele-operation drifts was much lower in comparison to other data types. Despite lots of practice, turning around a tight corner is extremely difficult, and requires a lot of precision. The sequence of events during data collection is as follows: We start at a random position, move towards our believed "starting optimal position", depicted as the blue dot \ref{fig: experiment-drifting-trajectory-graphic}, turbo the vehicle, perform the drift, slowly drive back, rinse, repeat. To see this setup in real-life, we have provided an image in the experimentation section. We also found that using a cone near the turning trajectory was very useful. \ref{fig:obst} 

\par
\begin{center}
\begin{figure}[h!]
  \centering
\includegraphics[width=.35\textwidth]{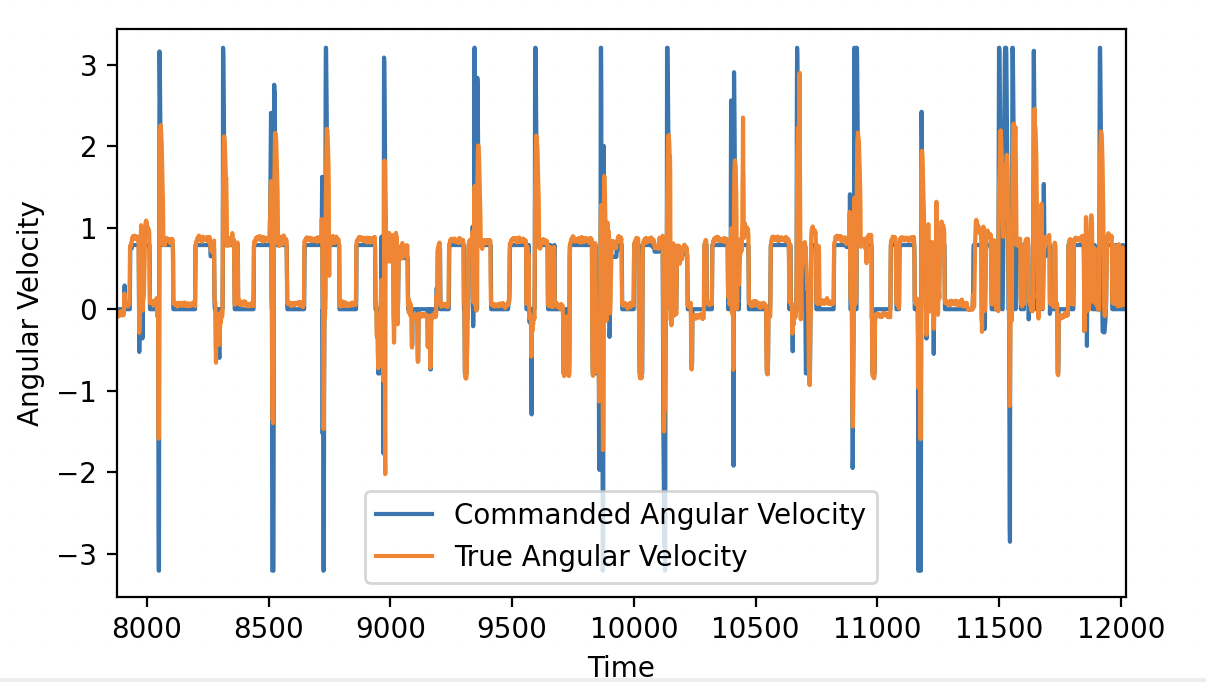}
\caption {Tight turning \#2, angular velocities overlaid.}
  \label{fig: extremely-tight-av-data}
\end{figure}
\end{center}

\begin{center}
\begin{figure}[h!]
  \centering
\includegraphics[width=.35\textwidth]{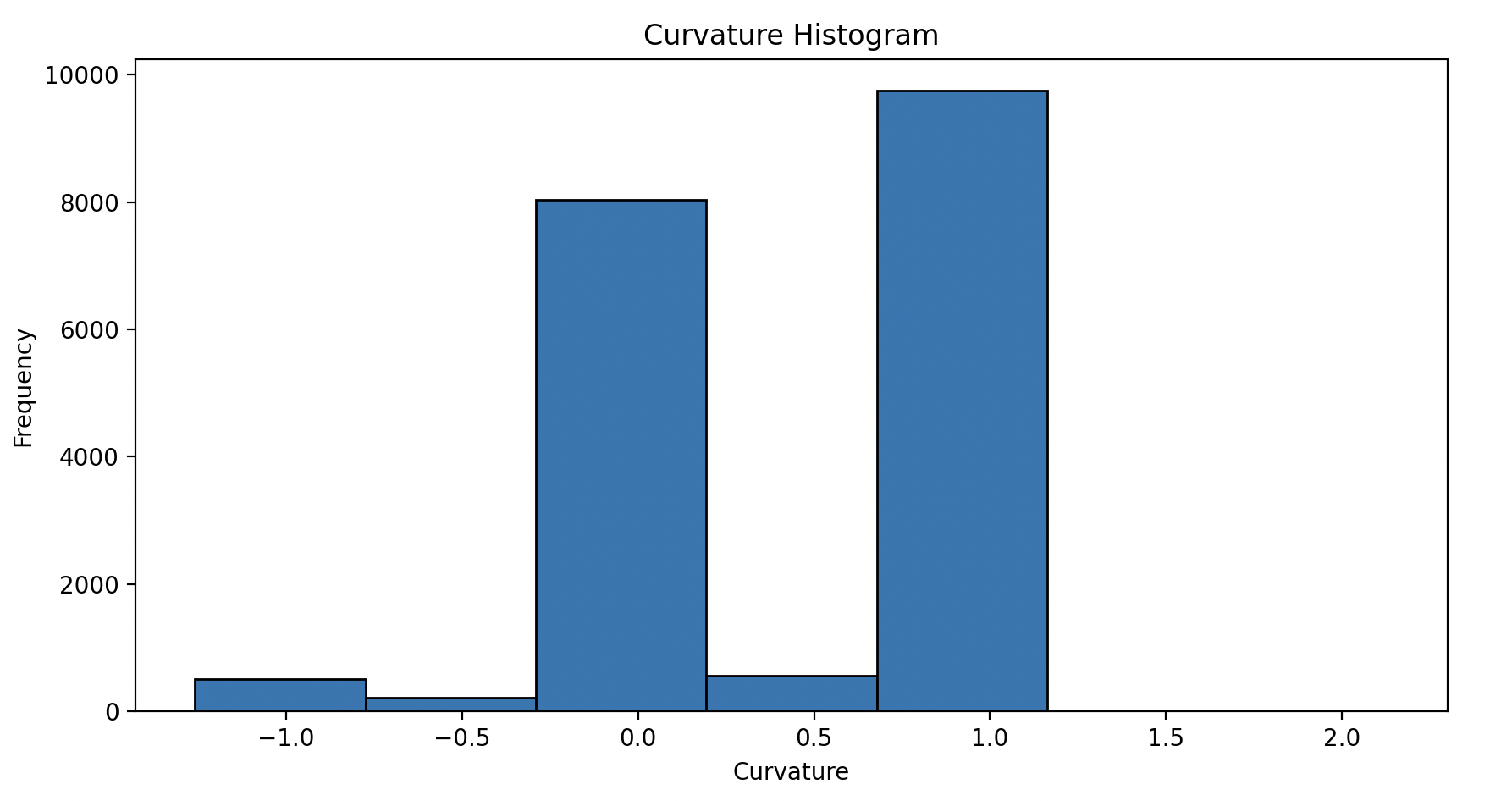}
\caption {Tight turning \#2, curvatures binned.}
  \label{fig: extremely-tight-curve-data}
\end{figure}
\end{center}

 We obtained some great trends from this data after plotting it \ref{fig: extremely-tight-av-data} \ref{fig: extremely-tight-curve-data}. We noticed that a great deal of our curvatures were extremely tight, taking majority frequency over the amount of linear movement during data collection. Additionally, the commanded velocity and angular velocity still had a quite clear difference, with more noticeable spikes as we intentionally slowed the car down between each drift to make the chart easier to read.

Our final step in the data collection process was to leverage the merging script discussed in the data extraction section to create our final training data for the model. After merging, we plotted the curvatures to get an insight into the diversity of our data. 

\begin{center}
\begin{figure}[h!]
  \centering
\includegraphics[width=.35\textwidth]{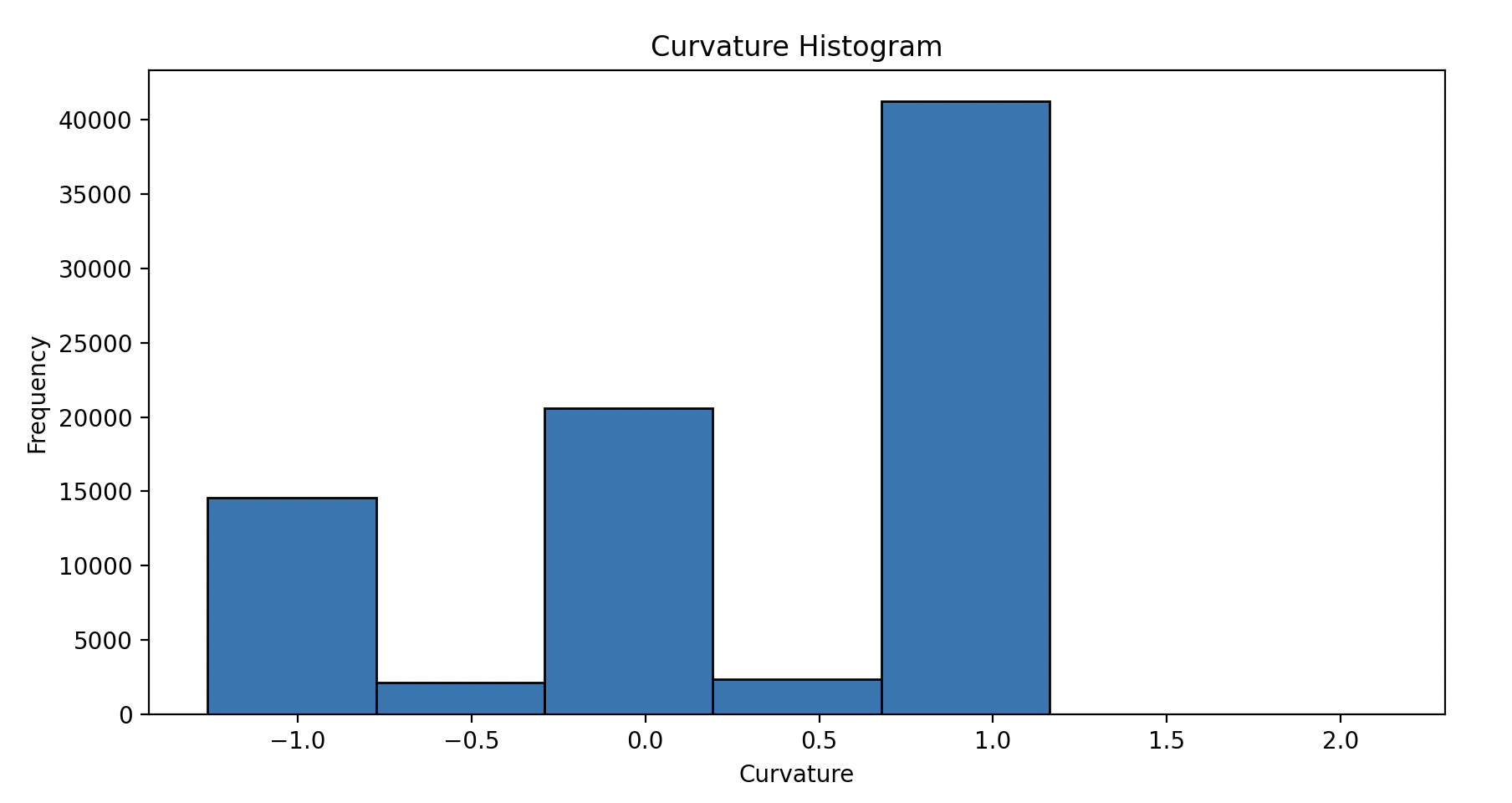}
\caption {Merged final training data, binned curvatures.}
  \label{fig: merged-curvature-binning}
\end{figure}
\end{center}

\begin{center}
\begin{figure}[h!]
  \centering
\includegraphics[width=.35\textwidth]{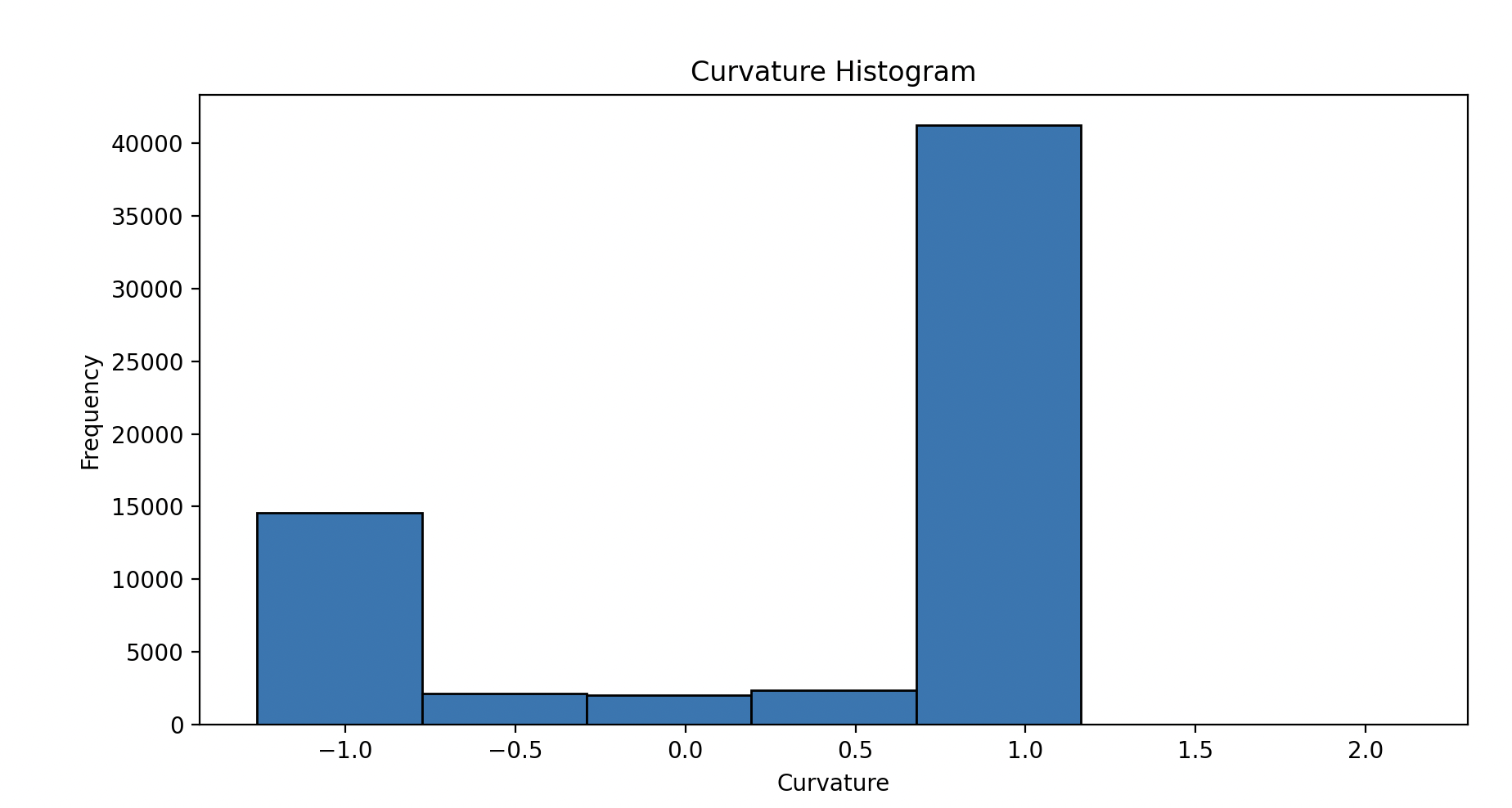}
\caption {Pruned 0.0 curvatures from final training data. This is the data used to train the neural network.}
  \label{fig: pruned-merged-curvature-binning}
\end{figure}
\end{center}

Referencing the curvature binning at figure \ref{fig: merged-curvature-binning}, it is clear that there is a lot of data with curvatures of zero. This is because on a straight path, the angular velocity of the vehicle is 0, which equates to a curvature value of 0. Being that our entire goal is to tighten turns and slips using the inverse kinodynamic model, we ultimately made the decision to write a script to prune out all the zero value curvatures in the training data, as that could potentially lead to improved model performance. In a way, we could argue that the frequency of zeros in our data could be equated to noise, and affect our network's ability to learn patterns robustly. Alternative upsides are that training speed would increase as well. After pruning out the training data, this was the resulting histogram \ref{fig: pruned-merged-curvature-binning}. At this point, we were ready to train our model.

\subsection{Problem Formulation}

\par Let us denote $x$ as the linear velocity of the joystick, $z$ as the angular velocity of the joystick, and $z'$ as the angular velocity measured off of the IMU unit on the vehicle. 

\par In the paper that inspired our work, the goal, generally, is to learn the function $f_{\theta}^{+}$ given the onboard inertial observations. More specifically, the paper formulates the function below: 

$$f_{\theta}^{+}(\Delta{x}, x, y) \approx f^{-1}(\Delta{x}, x, y)$$

\par We can denote $x$ as the linear velocity of the joystick, $z$ as the angular velocity of the joystick, and $z'$ as the angular velocity measured by the IMU unit on the vehicle. We will denote our desired control input as $u_{z}$. 

\par Our goal is to learn the function approximator $f_{\theta}^{+}$ based on the onboard inertial observations $z'$. $f_{\theta}^{+}$ then is used as our inverse kinodynamic model during test-time, in which it outputs our desired control input, $u_{z}$ to get us close to $z'$.
\begin{gather*}
f_{\theta}^{+}: (x, z') \rightarrow {NN} \rightarrow z\\
(x, z) \rightarrow f^{-1} \rightarrow u_{z}
\end{gather*}
\par At training time, we feed two inputs into our neural network architecture, which is joystick velocity and ground truth angular velocity from the IMU on the vehicle. The output of this model is the predicted joystick angular velocity. The learned model is our learned function approximator, which is then used as test time as the inverse kinodynamic model to give us our desired control, a corrected angular velocity for the joystick $u_{z}$ that gets us closer to the observation in the real world, $z'$.

\par Additionally, we leverage some basic mathematical formulations for graphing, inputting values into our model, etc.

\par Angular velocity is the rate in which an object rotates around an axis, in our case a vehicle. To calculate angular velocity, we can use velocity (aka. the speed along a curve) and the radius of the circular path to calculate the angular velocity, and work backwards in other regards as well. Here are some key formulations: 

$$av = \frac{v}{r}$$
$$r = \frac{1}{c}$$
$$av = v * c$$
$$c = av / v$$

\subsection{Model Architecture} \label{nn}
\par In the original IKD paper \cite{ikdhighsped}, they feed the IMU readings into a 600-dimensional vector, pass it through 256 neurons as a sequential auto-encoder and then feed it into a simple neural network representing the learned function approximator ($f_{\theta}^{+}$), as mentioned earlier in the paper. In the case of our training data, we only leverage the one reading from the IMU data, making the need of a sequential auto-encoder unnecessary. Below, we outline a description of our model, and show case a graphic of our architecture:

\par Since our extracted IMU reading is 1 dimensional, we feed this value directly into the $f_{\theta}^{+}$ alongside the joystick velocity value, making the input to our inverse kinodynamic model a 2 dimensional tensor. The model itself is a quite simple and compact neural network implemented in PyTorch. It consists of 3 fully connected Linear layers, with the first 2 layers having 32 hidden neurons. The correction layer outputs a 1 dimensional prediction of the joystick angular velocity. In our forward pass, we leverage the ReLU activation function, and avoid a sigmoid or tanh activation on the correction layer. This is because we do not normalize our values during training, and want to avoid our predicted angular velocity being bounded within 0 and 1 or -1 and 1. Since our formulation and data is quite simple (purely float values), we did not need to really worry about normalizing to see an improvement in training loss. Generally, it is important to note that angular velocity can range from -4 to 4, in the most extreme cases. 

\begin{figure}[h!]
  \centering
  \includegraphics[width=0.42\textwidth]{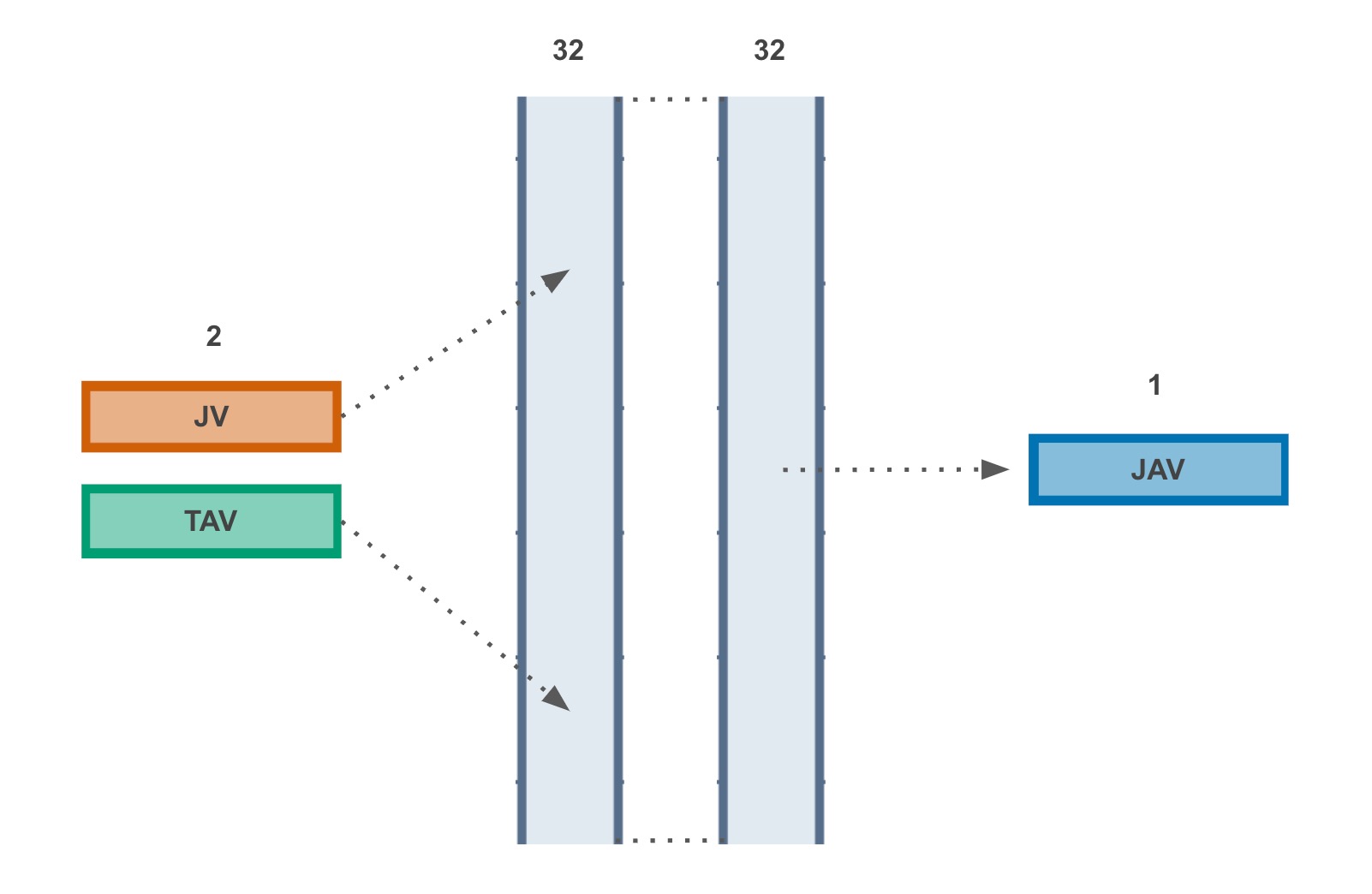}
  \caption{This is our neural network architecture. At train time, the input to the model is joystick velocity ($JV$) and ground truth angular velocity ($TAV$) from the IMU reading. The output of the model is the predicted joystick angular velocity ($JAV$). Thus, this becomes, the learned function approximator $f_{\theta}^{+}$ used as the inverse kinodynamic model during test-time.}
  \label{fig:nn}
\end{figure}

\subsection{Training} \label{training}

In this section, we discuss some basic information about the training of our neural network. To get more information regarding the architecture of the network, please refer to the respective section. 

\par
\begin{center}
\begin{figure}[h!]
  \centering
\includegraphics[width=.35\textwidth]{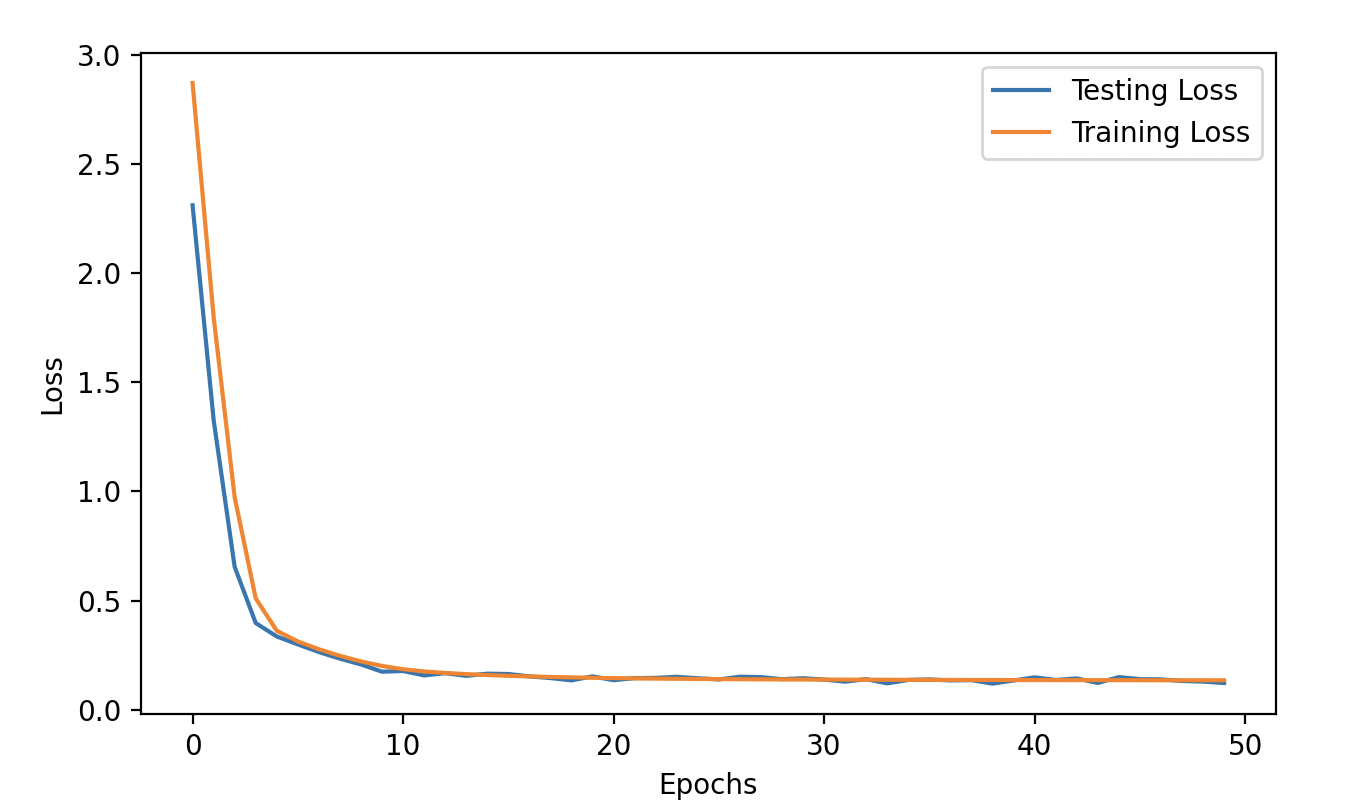}
\caption {Training and test loss of neural network plotted with respect to the number of epochs.}
  \label{fig: training-and-testing-loss}
\end{figure}
\end{center}

 Our neural network used a very standard training loop. We leveraged several emperical best practices from machine learning to stabilize training, including the AdamW optimizer, which a fixed batch size of 32. Training is performed using PyTorch on the M1 MacBook Pro CPU, and converges quite quickly, as the model size is small and the number of data points is minimal compared to modern deep learning research. Additionally, as discussed before, we avoid normalizing values as our data is simple float precision values and has a very well-defined range. We did not need any parallelization, train for 50 epochs, and leverage MSE (Mean Squared Error) loss function, which is a common empirical practice for simple neural networks. After performing training, we plotted both the training loss and testing loss, and found that the values were both decreasing and stabilizing around a particular value, which in normal machine learning can be a clear indication of an optimal fit. We show our resulting training and testing loss graph in the figure \ref{fig: training-and-testing-loss}.

\label{sec:method}

%% file: experiments.tex
\section{Experiments}
\label{sec:experiments}
In this section, we will discuss the experimental setup. 
\subsection{Environmental Setup}
First, we had to test various environments for the most clear slippage. As mentioned several times, the right surface, alongside the right teleoperated commands, are needed to execute the perfect drift. Thus, we tested various different surfaces around The University of Texas at Austin campus.

\begin{figure}[h!]
    \centering
    \includegraphics[width=.65\linewidth]{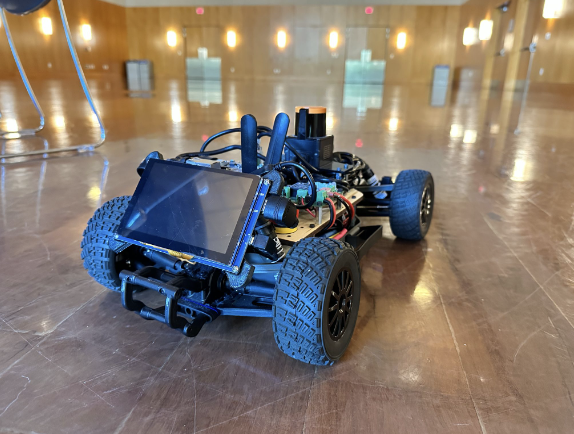}
    \caption{Floor of the SAC (Student Activity Center) ballroom; our first testing ground. Had a very nice and smooth surface, but we got kicked out (multiple times) for recording data here.}
    \label{fig:wcp}
\end{figure}
\par It can be seen in Fig. \ref{fig:wcp} that this testing ground had a very smooth and nicely polished floor, perfect for drifting our small-scale autonomous vehicle and collecting testing/training data. However, we got kicked out for attempting to test here as this room is by reservation only. 

\begin{figure}[h!]
    \centering
    \includegraphics[width=.65\linewidth]{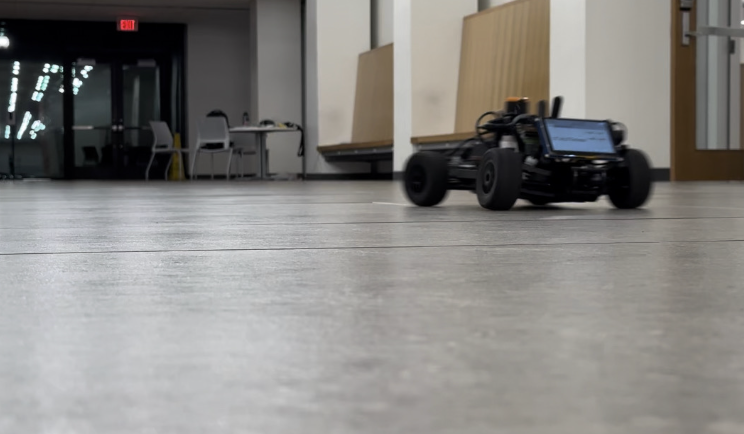}
    \caption{Floor of the WEL (Welch) building; great testing ground, but seldom empty. Usually not busy on weekends or late at night, but hard to collect data at these times as we don't have access to this building after hours.}
    \label{fig:welch}
\end{figure}

\par As seen in Fig. \ref{fig:welch}, the surface is also extremely smooth and not as grippy. Thus, we could easily drift here. However, other challenges, such as not having access to the building after hours, and routine daily traffic prevented us from being able to record training data here as well.

\begin{figure}[h!]
    \centering
    \includegraphics[width=.85\linewidth]{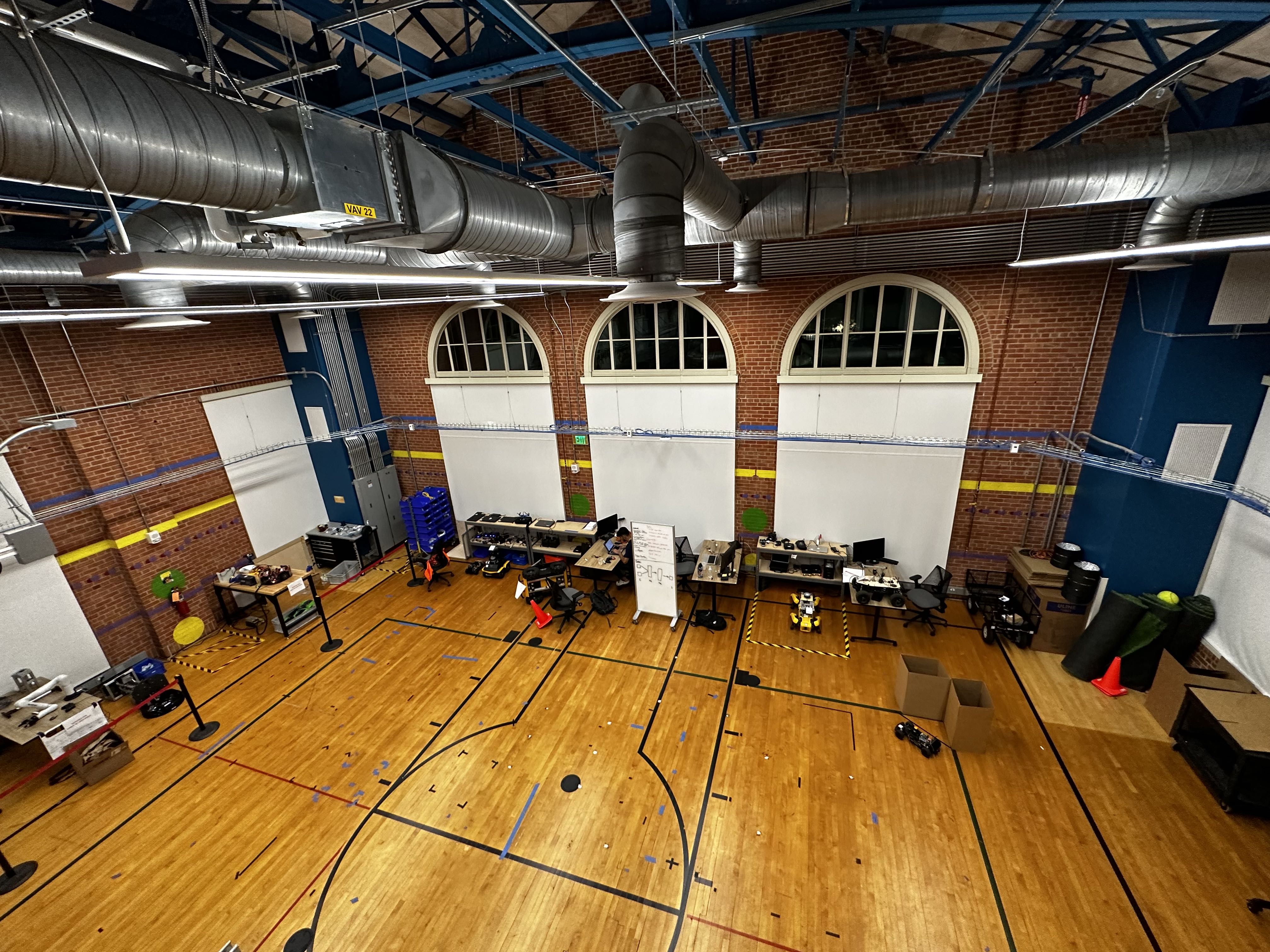}
    \caption{AHG (Anna Hiss Gymnasium) 2.202A, where all training and test data was collected.}
    \label{fig:ahg}
\end{figure}

\par Finally, we found the AHG (seen in Fig. \ref{fig:ahg}) as our testing/training ground for recording data. We found that the floor was perfect for executing drifts, tight turns, and even just collecting regular data for other training our IKD model.
\subsection{AHG Specifications and Limitations}

We'd just like to discuss the dimensions of the AHG that we used for testing our data, seen in the wooden floor in Fig. \ref{fig:ahg}. We used 35.41' x 21.67' as our testing area, performing all maneuvers in this space. 

\par Notably, this was a bit tight but it was the largest space we had access to; some limitations of this were that we were only able to collect tight turns and not as many loose turns as we would have liked to collect. That is evident in the training data that we collected, seen in Fig. \ref{fig: merged-curvature-turning}.

\par This lack of data in a fully diversified range was due to the limitations of the area that we were working within; we were not able to collect as much training data in this curvature range, as there was not much room to operate a vehicle at a high-enough speed and with a loose curvature. We noticed at such pairings of velocity and curvature, the car was quite likely to collide with objects in the environment; thus, our curvatures were quite limited. Generally, the takeaway idea was that when tele-operating the robot, you can only really turn at max curvatures, and turning with medium curvatures is difficult. 

\subsection{Simple Circle Testing with IKD Model}\label{circle}
Initially, we wanted to test our IKD model on a simple circle to ensure that our model actually learned off the training data, and was able to correct the curvature close to what was being commanded. Our testing setup was as follows: a commanded velocity of \textit{2.0 m/s} for the vehicle and a commanded curvature of \textit{0.7m}. 
\par For data collection, we simply drove the car around at a single velocity of \textit{2.0 m/s}, never applying the turbo speed. We executed many various curvatures and made sure that we recored various types of data to account for all curvatures that were possible in the testing space (more than 10 minutes of tele-operating the car).
\par
\begin{figure}[h!]
    \centering
    \includegraphics[width=.95\linewidth]{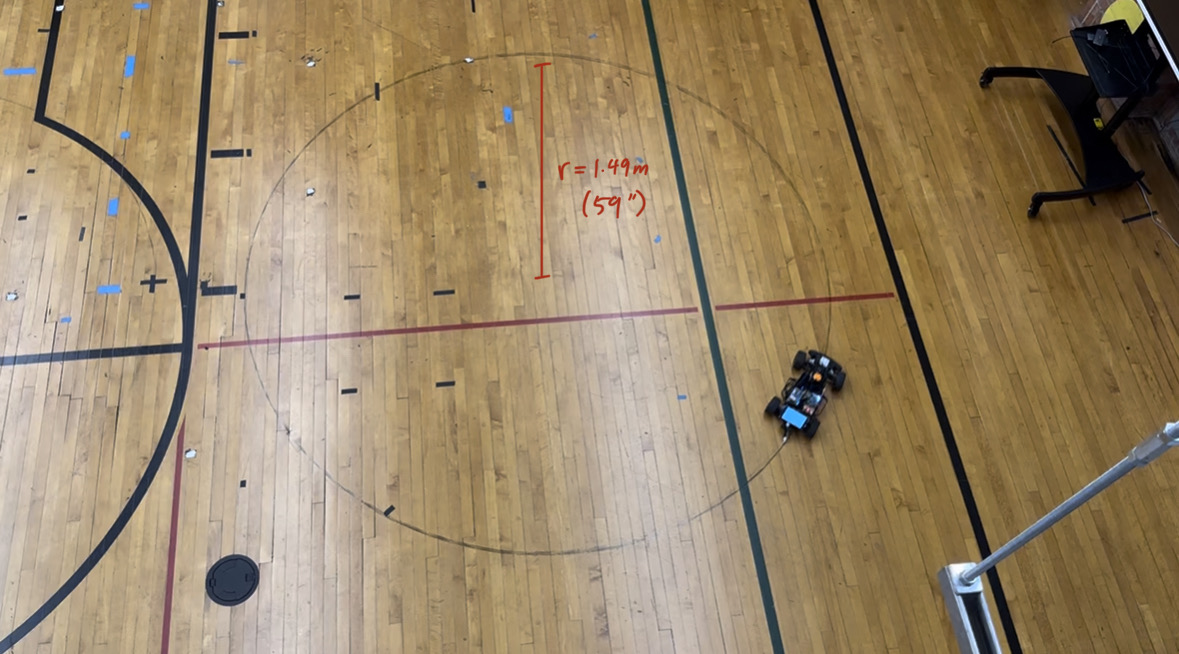}
    \caption{Non-IKD circle at commanded velocity of 2.0m/s and curvature of 0.7m. Radius measured was 1.49m (59").}
    \label{fig:withoutikdcircle}
\end{figure}

 Shown in Fig. \ref{fig:withoutikdcircle} is the actual circle that was executed on the car, with manually set curvatures and velocity in \verb|navigation.cc|. We attached a dry-erase marker to the rear-end of the vehicle to visually mark the trajectory, later erasing any marks left behind after our testing was complete.

It can be seen that despite the commanded curvature of 0.7m, the executed curvature was not the same. Calculating that value really quickly, with the observed measured radius of 1.49m:
\begin{gather*}
    \textit{curvature} = \frac{1}{\textit{radius}} \\ 
    \textit{cuvature} = \frac{1}{1.49m} \\ 
    \textit{curvature} = 0.667m
\end{gather*}

\par
\begin{figure}[h!]
    \centering
    \includegraphics[width=.95\linewidth]{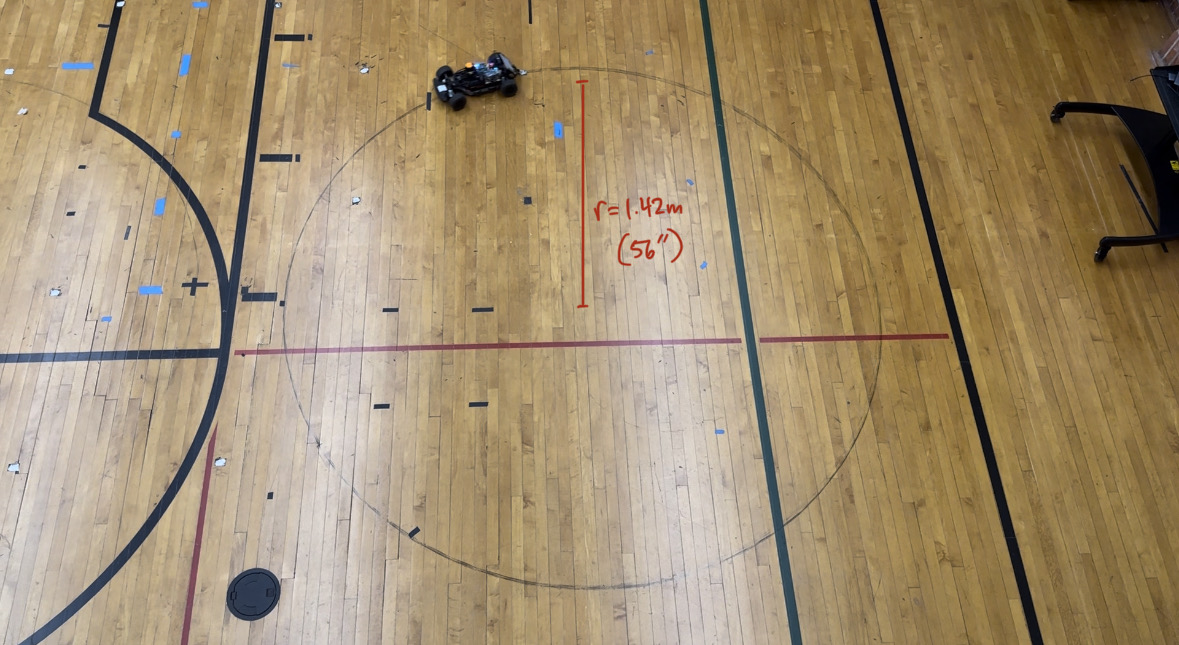}
    \caption{IKD-corrected circle at commanded velocity of 2.0m/s and curvature of 0.7m. Radius measured was 1.42m (56").}
    \label{fig:withikdcircle}
\end{figure}

 Now, after applying our IKD model, we noticed that the curvature changed. Take a look at Fig. \ref{fig:withikdcircle}.
\par It can be seen that the observed radius now changed to 1.42m. Calculating the new curvature using that value:
\begin{gather*}
    \textit{curvature} = \frac{1}{\textit{radius}} \\ 
    \textit{cuvature} = \frac{1}{1.42m} \\ 
    \textit{curvature} = 0.704m
\end{gather*}
\par This was evidence that our IKD model was learning off the trajectories that we fed it. Furthermore, we conducted an extension of this test for other curvatures. In experimentation, we tried 3 additional different curvatures; 0.12m, 0.63m, and 0.80m. Our velocity was kept standard at a rate of \textit{2.0 m/s}. We display the results in  Table \ref{tab:circletest}.
\begin{table}[H]
\setlength\tabcolsep{0pt}
\begin{tabular*}{\columnwidth}{@{\extracolsep{\fill}}lccc}
\toprule
\thead[l]{Commanded \\ Curvature} 
  & \thead{Executed\\Curvature} 
    & \thead{IKD-corrected\\Curvature} 
      & \thead{IKD-Commanded\\Deviation}\\
\midrule 
    0.12m        & 0.135m & 0.1172m & 2.33\%\\
    0.63m    & 0.634m & 0.6293m & 0.11\%\\
    0.80m & 0.810m & 0.8142m & 1.78\%\\
\bottomrule
\end{tabular*}
\caption{A chart of commanded, executed, and IKD-corrected curvatures using our IKD model. Note that velocity was kept standard at 2.0 m/s.}
\label{tab:circletest}
\end{table}

\par Notice that IKD was able to correct the executed curvature to more closely resemble the commanded curvature, with error values ranging from 0.11\% up to 2.33\%.

\subsection{Drifting Testing with IKD Model}
As proven in section \ref{circle}, our function approximator $f_{\theta}^{+}$ seemed to be working generally well given different commanded curvatures. This was indicative that it had learned the given terrain, as well as how the joystick velocity and angular velocity maps to the true angular velocity (from the IMU), and how to correct the angular velocity. We noticed that we needed diversified training data (couple minutes, lots of turns and twists) to truly learn the right curvatures and some limitations were definitely our space that we tested on.
\par Despite this, our next line of research was to apply our IKD model to a car that is autonomously drifting (using a teleoperated sequence of inputs). Initially, the challenge was understanding how to run such a random sequence of inputs for drifting autonomously. To alleviate this, we created a file (in \cite{autogit}) called \verb|csv_to_circular_linked_list.cpp| that converts records from the CSV into a circular linked list to be read in. As mentioned previously, the car's navigation commands are run at a frequency of \textit{20 Hz}, so by feeding it in the circular linked list commands and just passing the pointer to the next node at each function call, we are able to read off the recorded values for a teleoperated drift and replicate it. We noticed that sometimes, even on the same terrain, the behavior was not entirely the same; often, small changes in the world (bumps in the ground floor) accounted for a different drift, with the car still ending up in roughly the same space.
\begin{figure}[H]
    \centering
    \includegraphics[width=.75\linewidth]{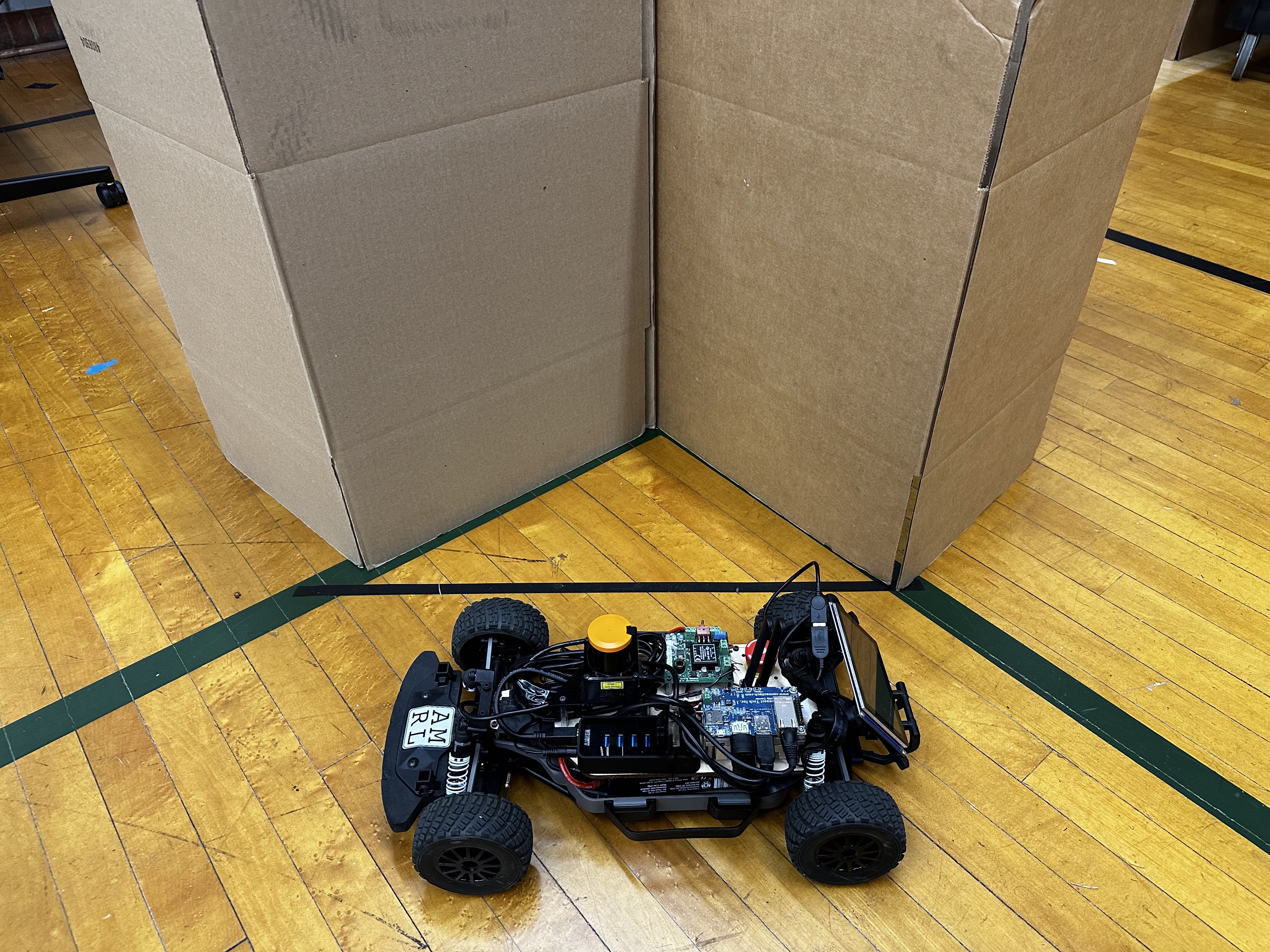}
    \caption{Drifting setup, with two cardboard boxes as "obstacles".}
    \label{fig:obst}
\end{figure}
\par Our testing setup for drifting can be seen in \ref{fig:obst}; we set up two cardboard boxes (to represent obstacles) in the corner of the AHG, ensuring that the black tape width between these boxes was just enough to be the width of the car (19"). Our idea here is that on the initial drifting data set that we collected, we noticed that the car collided with these two obstacles. 
\par Using the IKD model and the function approximator $f_{\theta}^{+}$, we should be able to correct the slippage to be a little tighter, thus avoiding the area of colliding with the boxes. This idea can further be extended to real vehicles; can a car drift in some autonomous sequence, avoiding nearby obstacles by correcting the slip if necessary?

\par Collecting data for drifting wasn't as easy as we thought. We first had to learn how to control the car using a teleoperated sequence of joystick events, especially on how to feed that into the existing ROS nodes onboard the UT AUTOmata car itself. We settled on using our fresh copy of \verb|cs378_starter| code that we downloaded earlier, and began by recording a bag file of just \verb|/joystick| topic information. To perform a teleoperated sequence of events it is important to only collect relevant topic information, and the joystick was all that we needed.
\par From here, we took that bag file, and converted this into a CSV file. We used dummy IMU data to align the data, and ensure that we were getting the correct sequence of events being extracted from the joystick, and then created a Python script (\verb|get_joystick_data.py|) to extract only the joystick data to a \verb|.txt| file, where each row was a tuple of \verb|[v,av]|. Thus, by extracting this data, it is ready to be fed and decoupled into velocity and curvature pairs in the actual codebase on the car.
\par After this extraction was complete and we had a list of the joystick data in a text file, we had to find the relevant areas to plug this in on the car. Through some debugging, we found that we can feed in a velocity and a curvature into a message for the \verb|drive| node, and we created a linked list that read from the text file and stored all joystick and angular velocity pairs as a node. This was all done in the constructor in \verb|navigation.cc|; from here, we can use the \verb|run()| function that we talked about earlier.

\par The \verb|run()| function is called at a frequency of \textit{20 Hz}, so having this was a bit of a challenge as we had to figure out how to put a sequence of events to be run instantaneously. Instead, we settled on creating a global pointer called \verb|line_num| that tracks the current line and only extracts information from the node \verb|line_num| is pointing to; it then publishes this information to the drive message, which is published onto the actual \verb|drive node|.

\par This is essentially our testing framework; now, given any CSV file that we have recorded, we can extract only the joystick data to a text file, and then load it onto the car to be run autonomously. Thus, the framework for autonomous drifting is ready. The charts for all data collected can be found in our GitHub \cite{autogit}.

\subsubsection{Loose Drifting, Counter-clockwise and Clockwise}
Our first test was to see if we could drift the car using the IKD-corrected model in a loose curvature. We used the testing setup discussed in \ref{fig: loose-drifting-diagram}, where we had two cones that were a distance of 84 inches apart. 
\par We began by recording a bagfile of the teleoperated inputs, and then converted this into the text file talked about earlier in the paper. We conducted a total of four trials for each loose drift, two in the counter-clockwise and two in clockwise direction.
\par From here, we fed this teleoperated sequence of events from the joystick into the actual model that we trained earlier, and had our IKD model loaded on the fly. Thus, we can make predictions at runtime and see what is outputted on the actual car by empirically observing it. We conducted trials in which we watched the car drift autonomously both with and without IKD, and in the clockwise and counter-clockwise directions. Our results are show in table \ref{tab:loose}.

\begin{table}[H]
\setlength\tabcolsep{0pt}
\begin{tabular*}{\columnwidth}{@{\extracolsep{\fill}}lcc}
\toprule
\thead[l]{Drifting Direction} 
  & \thead{Tightened Turn \\ Rate} 
    & \thead{IKD Corrected Rate} \\
\midrule 
    Clockwise      & 50\% & Non-noticeable\\
    Counter-clockwise   & 100\% & Noticeable\\
\bottomrule
\end{tabular*}
\caption{A chart of drifting in a loose fashion, with clockwise and counter-clockwise directions. Trials done were 2 for each direction, with empirical visual observations for IKD corrections.}
\label{tab:loose}
\end{table}
 As seen above, we noticed that in the counter-clockwise direction that the IKD model actually tightened our turns quite well. Despite going slowly and loosely around the cones, the IKD model was able to tighten this and almost even hit the cone in a few trials, clearly showing that IKD was preventing slip on these turns. This was pretty impressive.
\subsubsection{Tight Drifting, Counter-clockwise and Clockwise}

For our second test, we put the car through a very complex scenario, which was discussed earlier and can be seen in the graphic in the data collection section. We used our script to autonomously record the drift, and ensured that the car could clear it. But.. when we applied our IKD model to the same sequence of controls, the result was a failure rate on every trial. The vehicle would perform a drift, but would shorten the distance and tighten the trajectory, not clear the turn completely. We discuss this later in the paper, but assume that this is likely happening due to the limited amount of drifting data at such extreme trajectories, making the predictions of the model less robust in comparison to the loose drifting predictions.




%% file: discussion.tex
\section{Challenges and Future Direction}

\par In this paper, we explored a formulation of the inverse kinodynamic model to correct the control input of angular velocity on a tight turn to perform an optimal drift at different trajectories. While interesting, the problem itself was quite complex, and required a lot of moving parts. Our minor success with loose drift likely stemmed from the minimal correction required by the IKD model to tighten the drift. Additionally, with high slippage, as discussed in the paper, the difference between the true angular velocity and the commanded angular velocity is quite large. Our insight is that the model did not have a good enough data set for both lateral drifting positions. Additionally, our model did converge quite quickly, likely due to the batch size being used on a data-set of only 66,000 points. We also believe that our environment was one of the key factors that set back a full-scale success for the model. The AHG dimensions were quite limited, but unfortunately was the most reasonable and accessible space to do continuous and effective work on the vehicle. We also believe that the formulation of IKD that we used was too simple, and the general performance of the drifting could have been improved if we had used real sense data as a part of the learning formulation. We don't believe that the size of our network really had an impact, as in the original paper, the learned function approximator was quite trivial and effective.

\label{sec:discussion}

%% file: conclusion.tex
\section{Conclusion}

\par In this paper, we proposed a modified version of Inverse Kinodynamic Learning for safe slippage and tight turning in autonomous drifting. We show that the model is effective for loose drifting trajectories. However, we also find that tight trajectories hinder the models performance and the vehicle undershoots the trajectory during test time. We demonstrate that data evaluation is an essential part of learning an inverse kinodynamic function, and that the architecture necessary to have success is simple and effective. 

\par This work has the potential of becoming a stepping stone in finding the most effective and simple ways to autonomously drift in a life-or-death situation. Future work should focus on collecting more robust data, using more inertial readings and sensor readings (such as real-sense, other axes, or LiDAR). We have open-sourced this entire project as a stepping stone in these endeavors, and hope to explore our ideas further beyond this paper \cite{autogit}.

%% file: acknowledgement.tex
\section{Acknowledgements}

Special thanks to Pranav, Rahul and Arnav (of the UT AMRL Laboratory) for helping us overcome the learning curve to accomplish the completion of this paper. Additionally, we'd like to give a special thanks to Dr. Joydeep Biswas for providing us the resources and access necessary to complete this work. All resources used were a part of the UT Autonomous Mobile Robotics Laboratory (AMRL), led by professor Joydeep Biswas. 